\documentclass[letterpaper]{article} 
\usepackage{aaai25}  
\usepackage{times}  
\usepackage{helvet}  
\usepackage{courier}  
\usepackage[hyphens]{url}  
\usepackage{graphicx} 
\usepackage{natbib}  
\usepackage{caption} 
\usepackage{algorithm}
\usepackage{algorithmic}

\usepackage{newfloat}
\usepackage{listings}
\DeclareCaptionStyle{ruled}{labelfont=normalfont,labelsep=colon,strut=off} 
\floatstyle{ruled}
\newfloat{listing}{tb}{lst}{}
\floatname{listing}{Listing}

\def\eg{{\em e.g.,~}}

\def\ie{{\em i.e.,~}}

\usepackage{subcaption}
\usepackage{pifont}

\newcommand{\cmark}{\ding{51}}%
\newcommand{\xmark}{\ding{55}}%

\usepackage{makecell}
\usepackage{multirow}

\newcommand{\DATASET}{VDAct}
\newcommand{\METRIC}{VDEval}

\newcolumntype{K}[1]{>{\centering\arraybackslash}p{#1}}

\title{A Video-grounded Dialogue Dataset and Metric for Event-driven Activities}
\author{
    Wiradee Imrattanatrai\textsuperscript{\rm 1}\equalcontrib,
    Masaki Asada\textsuperscript{\rm 1}\equalcontrib,
    Kimihiro Hasegawa\textsuperscript{\rm 2}, \\
    Zhi-Qi Cheng\textsuperscript{\rm 2},
    Ken Fukuda\textsuperscript{\rm 1},
    Teruko Mitamura\textsuperscript{\rm 2}
}
\affiliations{
    \textsuperscript{\rm 1}National Institute of Advanced Industrial Science and Technology (AIST)\\
    \textsuperscript{\rm 2}Language Technologies Institute, Carnegie Mellon University\\

    wiradee.imrattanatrai@aist.go.jp, masaki.asada@aist.go.jp
}

\usepackage{bibentry}

\begin{document}

\maketitle

\begin{abstract}
This paper presents VDAct, a dataset for a Video-grounded Dialogue on Event-driven Activities, alongside VDEval, a session-based context evaluation metric specially designed for the task. 
Unlike existing datasets, VDAct includes longer and more complex video sequences that depict a variety of event-driven activities that require advanced contextual understanding for accurate response generation. 
The dataset comprises 3,000 dialogues with over 30,000 question-and-answer pairs, derived from 1,000 videos with diverse activity scenarios. 
VDAct displays a notably challenging characteristic due to its broad spectrum of activity scenarios and wide range of question types. 
Empirical studies on state-of-the-art vision foundation models highlight their limitations in addressing certain question types on our dataset. 
Furthermore, VDEval, which integrates dialogue session history and video content summaries extracted from our supplementary Knowledge Graphs to evaluate individual responses, demonstrates a significantly higher correlation with human assessments on the VDAct dataset than existing evaluation metrics that rely solely on the context of single dialogue turns.
\end{abstract}


\begin{links}
    \link{Resources}{https://github.com/aistairc/VDAct}
\end{links}

\section{Introduction}
\label{sq:introduction}
\begin{figure*}
    \centering
    \includegraphics[width=\linewidth]{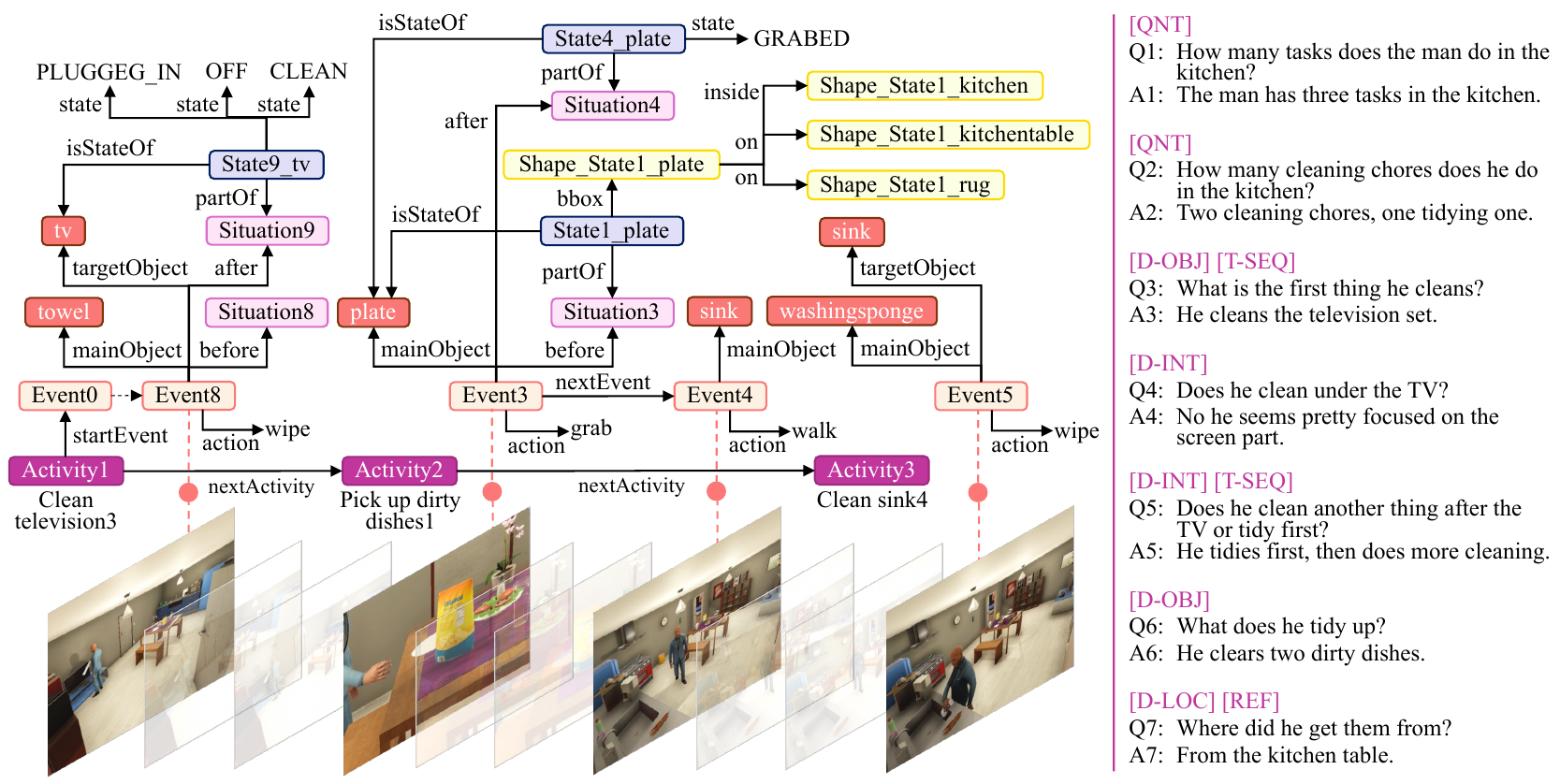}
    \caption{VDAct with example dialogue (turns 1-7) for an activity scenario video with corresponding KG elements for the events. Each input utterance (\ie question) is labeled with the relevant question types. Note that the object IDs are omitted from the KG illustration.} 
    \label{fig:dataset}
\end{figure*}

\begin{table*}[t]
\fontsize{8}{10}\selectfont
\begin{tabular}{p{0.12\linewidth} | c | p{0.73\linewidth}}
\hline
\multicolumn{2}{c|}{Category} & \multicolumn{1}{c}{Description} \\ \hline
\multicolumn{1}{p{0.12\linewidth}}{\textbf{(1) Descriptive}} \\ 
- Agents & D-AGT & Questions about characteristics or states of the agents \\
- Actions & D-ACT & General questions related to actions without specifying objects  \\
- Objects & D-OBJ & Questions about objects involved in actions, those upon which actions are performed, or the states of objects \\
- Interactions & D-INT & Questions about whether actions were performed on specific objects \\
- Locations & D-LOC & Questions concerning the whereabouts of agents, objects, or where the actions were performed \\ \hline
\multicolumn{1}{p{0.12\linewidth}}{\textbf{(2) Temporal}} \\ 
- Sequence & T-SEQ  & Questions related to the temporal sequence of actions w/wo objects or questions related to specific points in time \\
- Frequency & T-FRQ & Questions about the frequency or duration of actions performed w/wo objects \\ \hline
\textbf{(3) Explanatory} & EXP & Questions seeking explanations for how and why actions were carried out  \\ \hline
\textbf{(4) Quantitative} & QNT & Questions related to quantity or number of agents or objects \\ \hline
\multicolumn{1}{p{0.12\linewidth}}{\textbf{(5) Other}} \\
- Reference & REF & Questions that reference previous dialogue turns or require answers based on the dialogue history \\ 
- Supplementary & SUP & Questions seeking further details or information about the activities \\
- Video Attributes & VID & Questions about attributes such as audio, quality, length of the video, or the language spoken within the video \\
- Opinions & OPI & Questions asking opinions or facts about the agents, actions, and objects that are subjective or cannot be verified \\ \hline
\end{tabular}
\caption{Categories and types of questions involve in VDAct.}
\label{tab:question-type-definition}
\end{table*}

The video-grounded dialogue task involves generating responses to user utterances based on the video content. This task poses significant challenges particularly when dealing with videos presenting compound activities where multiple related events occur in sequence. With these event-driven activities, advanced system capabilities in multimodal understanding, temporal reasoning, and contextual interpretation are required to align visual cues with conversational context and handle dynamic changes in the video. Although several datasets exist for video-based reasoning through question answering, only a few benchmark datasets are available for the video-grounded dialogue task~\cite{alamri2019audio, pasunuru-bansal-2018-game}. These datasets mostly feature short videos depicting simple activities and a limited range of question types, while in real-world scenarios, dialogue discussions often center around multifaceted activities about how various events with different associated actions are temporally and contextually related. Thus, exposure to dialogues on event-driven activities would expand the system's ability to handle complex interactions and improve its capacity to generate accurate and relevant responses.




Thus, to advance research and development of video-grounded dialogue systems on event-driven activities, we introduce a new dataset named ``\textbf{V}ideo-grounded \textbf{D}ialogue on Event-driven \textbf{Act}ivities'' (\DATASET). 
This dataset includes dialogues based on daily scenarios where each involves multiple activities with long sequences of events. We opted to utilize virtual simulation videos that allow a variety of activity combinations. Unlike existing datasets which primarily focus on descriptive questions, VDAct includes several other categories to capture a broader range of interactions as shown in Table \ref{tab:question-type-definition}. In addition to the \textbf{descriptive} questions which aim to obtain factual information about the activities, we incorporate three other main categories including \textbf{temporal} questions, which focus on temporal aspects, such as timing, duration, and sequence of events or activities; \textbf{explanatory} questions which explore the reasons or causes behind events or activities; and \textbf{quantitative} questions, which seeks numerical or quantitative data. Additionally, VDAct incorporates questions related to the video and dialogue attributes, as well as open-ended and subjective questions.

In addition to videos and dialogues, we enrich our dataset with Knowledge Graphs (KGs) as supplementary information. Given that our target videos represent compound activities involving multiple events, KGs could be useful for both system development and evaluation as they offer detailed information that links visual cues to the structured information. This information includes event sequences, transitions between events, action-object interactions, and changes in agent and object states, as illustrated in Figure \ref{fig:dataset}.

Furthermore, existing evaluation metrics, such as those designed for text generation~\cite{papineni2002bleu, banerjee2005meteor, lin2004rouge}, and QA tasks~\cite{manas2024improving, chan-etal-2023-clair, wada2024polos}, are insufficient for assessing the quality of generated responses in the context of dialogues. 
These metrics usually assess system-generated text by comparing it to a reference text, but they often overlook cases where multiple responses can be equally valid. For example, in response to Q7 in Figure \ref{fig:dataset}, both ``He grabbed the plates from the kitchen table.'' and ``They were on the rug, which was on top of the kitchen table.'' are valid when considering additional context from dialogue history and KGs. Although neither Q7 nor A7 mentions what objects they are referring to, the first response is deemed correct because it aligns with prior dialogue turns. Additionally, the KG information for the related event (\ie Event3 of Activity2) shows that the plate was on the rug, making the second response also accurate. However, because the existing metrics only compare responses to the reference and do not consider this additional context, they cannot fully capture the correctness of both responses and as a result, evaluate them as partially or completely inaccurate.
Thus, we propose a new LLM-based evaluation metric that integrates \textit{session-based} context including a summary of KG information and dialogue history, rather than relying on \textit{turn-based} context. This inclusion allows the evaluation model to verify the responses more comprehensively.

To this end, our main contributions are three-fold. First, we introduce a new video-grounded dialogue dataset focused on event-driven activities and provide accompanying KGs for the activity scenarios. Second, we propose a new evaluation metric, specifically for the video-grounded dialogue task, to address the unique aspects of dialogues where multiple responses are possible and inference from summary and dialogue history are required. Finally, we selected Vision LMs (VLMs) to evaluate their performance on our newly constructed dataset and evaluate the correlation between the results and our new proposed metric.

\section{Related Works}
This section reviews the existing datasets and the existing evaluation metrics for the video-grounded dialogue task.
\paragraph{Video-grounded Dialogue Datasets}
In terms of video understanding, several video-grounded QA datasets have been introduced. Among them, a few provide KGs such as STAR~\cite{wu2021star_situated_reasoning} which presents multi-choice QA samples along with situation hypergraphs based on real-world videos, and EgoTaskQA~\cite{jia2022egotaskqa} which introduced a QA dataset with annotations of object status, human-object and multi-agent relationships, and causal dependency structures between actions, all derived from egocentric videos. However, while QA tasks treat each question independently, dialogue tasks build on prior interactions of questions and answers. This offers a key advantage for system development as the sequential nature of dialogue enables the system to maintain continuity and deliver context-aware responses by referencing previous turns. For the video-grounded dialogue datasets, VisDial~\cite{das2017visual} treated the problem of visual dialogue as a multi-turn QA where the system is expected to answer questions given dialogue history and corresponding images. Audio Visual Scene-aware Dialog (AVSD)~\cite{alamri2019audio} extended the work from VisDial to include additional modalities including videos with audio signals. Twitch-FIFA~\cite{pasunuru-bansal-2018-game} introduced a video-context dialogue dataset based on live-broadcast soccer games and chat from Twitch.tv. Video-grounded Scene and Topic AwaRe dialogue (VSTAR)~\cite{wang-etal-2023-vstar} introduced a large-scale benchmark dataset for understanding the dialogue between characters in a TV series.
These datasets usually contain short videos with simple activities and focus mainly on descriptive questions. This constraint limits the system's ability to learn from more complex scenarios that involve multiple related events occurring in sequence.
Thus, we propose a new dataset with longer videos depicting complex activities, and dialogues presenting a variety of question types.

\paragraph{Evaluation Metrics for Video-grounded Dialgoue}
Developing an effective evaluation metric for the video-grounded dialogue task presents a significant challenge. Previous studies reported that classic evaluation metrics such as BLEU, METEOR, and ROUGE showed low correlation with human evaluations for the video-grounded dialogue task~\cite{liu-etal-2016-evaluate, alamri2019audio}. Considering this point, AVSD employs ranking-based evaluation metrics and a discriminative ranking task setting where the model prediction is selected from candidate answers. 

Learning-based metrics such as PAC-S~\cite{sarto2023positive} and Polos~\cite{wada2024polos} have been proposed and showed a high correlation with human evaluation for the image captioning task. However, these metrics require vast amounts of human scores to train the metrics. This makes it difficult to apply the metrics on the video modality.
LLM-based evaluation metrics, such as CLAIR~\cite{chan-etal-2023-clair} for image captioning, and LAVE~\cite{manas2024improving} and LLM-Acc/Rel~\cite{maaz2024videochatgpt} for VQA have been gaining attention in recent years. However, these metrics only compare the generated text with the reference of a single QA turn without considering additional contextual information that is beneficial for the evaluation.

In this study, we exploit the advantage of the video context being directly linked to event-centric structured KGs and dialogue history to propose a new LLM-based evaluation metric for the video-grounded dialogue task.


\section{\DATASET{} Dataset}
\subsection{Data Collection}

We prepare target scenario videos for dialogue generation, followed by employing crowdsourced workers to create dialogues discussing the event-driven activities depicted in the videos. To support the video-grounded dialogue task, we additionally include scenario KGs and their summaries. The following subsections detail the collection process for each data component.

\subsubsection{Preparation of Scenario Videos}

To gather video data representing daily living activities, we leverage the VirtualHome2KG dataset~\cite{egami2023synthesizing}\footnote{\url{https://github.com/KnowledgeGraphJapan/KGRC-RDF/tree/kgrc4si}}, which integrates KGs with video data for tasks such as activity recognition. VirtualHome2KG contains simulation videos depicting various daily activities performed by a single agent in a 3D virtual space using the VirtualHome platform~\cite{puig2018virtualhome}. Each video captures an activity in a unique home environment, varying in room layouts and camera angles. VirtualHome2KG relies on activities of 11 classes defined by HomeOntology~\cite{vassiliades2020knowledge} such as BedTimeSleep, EatingDrinking, FoodPreparation, HouseArrangement, and others, as well as one additional class, Abnormal. Each activity is associated with a program representing a sequence of events involving actions and objects. For example, an event of ``[WALK] $\langle$television$\rangle$ (297)'' where the number denotes the object ID. The dataset consists of 3,530 videos covering 706 activities across 12 categories, 7 environment setups, and 6 viewpoints, including an indoor camera switching view, character rear views, and fixed camera angles.


As the target for our dialogue is to analyze the daily scenario consisting of multiple activities, we combined 2-5 available activities within VirtualHome2KG as an activity scenario. 
Following are the constraints we imposed to obtain high-quality videos. First, we need to ensure the seamless connectivity of the activity videos as scenario videos. Thus, we chose to combine the activities where each occurs in the same environment setup and that the agent starts and ends the activity in the same room as its previous and succeeding activity, respectively. As there can be too many possible activity combinations, we do not allow the same activity to be in the same scenario, as well as limit the scenario video to 1 to 5 minutes of the combined activity videos.
Out of six viewpoints, we selected activity videos from either an indoor camera switching view or a fixed view at a room corner as they show the clearest depiction of activities for a scenario.

With a set of candidate scenarios, we further filtered out scenarios with some criteria to diversify the activity combination. Firstly, we set the limit for the number of individual activities appearing as the first activity as 8. Secondly, we excluded scenarios that have more than half of their activities duplicated across other scenarios. 
Lastly, we limit each adjacent activity pair to appear in no more than two scenarios. Thus, we obtained 3,021 unique activity combinations as scenarios. From this list, we randomly sample 1,000 scenarios for preparing the dialogues.

\subsubsection{Dialogue Data Creation}
For the creation of dialogue data, we hired six crowdsourced workers through a reputable third-party company specialized in creating and collecting language data for NLP research and development. The goal of this data collection step is to obtain dialogue sessions demonstrating the information exchange of the person's activity scenario between a pair of annotators. For each given scenario, the two annotators were assigned different roles to have a formal discussion about the scenario. One annotator was assigned to act as the investigator with a responsibility to investigate the person's behavior and figure out how the person performs the activities by asking questions to the corresponding annotator. The investigator was not allowed to watch the videos but was given an unordered list of activities to provide some ideas about the scenario. The other annotator, acting as the correspondent, was given a list of activities, and was assigned to watch the videos to provide accurate answers to the investigator.

To cover multiple types of questions, we chose to provide examples from four main categories (\ie descriptive, temporal, explanatory, and quantitative) that can be incorporated into the dialogues. Moreover, we also provided a few example dialogues to the annotators to have a clear picture of the task. We do not set strict minimum or maximum limits on the number of each question type per dialogue, as such a limitation could negatively impact the natural flow of the conversation. Instead, we instructed annotators to include as many question types within each dialogue to ensure type coverage.


For 1,000 scenarios from the previous data collection step, we formed three different annotator pairs. In half of the scenarios, one annotator was the investigator, while the other was the correspondent. The role of the pair switches for the latter half of the scenarios. This approach ensured that each annotator had the opportunity to play both roles, potentially leading to more diverse and comprehensive dialogues. 
Moreover, to ensure that the annotators fully understand the task instruction, we manually reviewed the initial dialogues created by each annotator pair before allowing them to proceed with the rest of the process. In total, we obtained 3,000 distinct dialogues with over 30,000 turns (\ie question-answer pairs) through this process. 

\begin{table}[t]
\fontsize{7.5}{10}\selectfont
    \centering
    \begin{tabular}{lp{0.4\linewidth}}
    \hline
Triplet & Sentence \\\hline
\makecell[l]{(event0, \textit{from}, bedroom75) \\ (event0, \textit{action}, walk) \\ (event0, \textit{mainObject}, bathroom11)} & \makecell[l]{The person is in the bedroom. \\ He walks to the bathroom.} \\ \hline
\makecell[l]{(door53, \textit{inside}, bathroom) \\ (door53, \textit{state}, OPEN) \\ (event1, \textit{action}, walk) \\ (event1, \textit{mainObject}, door53)} & \makecell[l]{The door is inside the bathroom. \\ The door is OPEN. \\ He walks to the door.} \\ \hline
\makecell[l]{(event2, \textit{action}, close) \\ (event2, \textit{mainObject}, door53)} & \makecell[l]{He closes the door.} \\ \hline
\makecell[l]{(door53, \textit{state}, CLOSED) \\ (toilet, \textit{inside}, bathroom11) \\ (event3, \textit{action}, walk) \\ (event3, \textit{mainObject}, toilet46)} & \makecell[l]{The door is CLOSED. \\ The toilet is inside the bathroom. \\ He walks to the toilet.} \\ \hline
\makecell[l]{(toilet46, \textit{close}, character1) \\ (character1, \textit{inside}, bathroom11) \\ (event4, \textit{action}, sit) \\ (event4, \textit{mainObject}, toilet46)} & \makecell[l]{The toilet is next to the person. \\ The person is inside the bathroom. \\ He sits on the toilet.} \\ \hline
    \end{tabular}
    \caption{Example triplets and the corresponding sentences for template-based video summaries from KGs.}
    \label{tab:kg2text-template}
\end{table}

\subsubsection{Knowledge Graph Collection for Scenarios}
As supplementary information for the dataset, we collected KGs for scenarios by relying on KGs for activities provided by VirtualHome2KG. Each activity KG consists of 9 main node types including ${\rm Activity}$, ${\rm Event}$, ${\rm Action}$, ${\rm Situation}$, ${\rm Object}$, ${\rm State}$, ${\rm Attribute}$, ${\rm StateVal}$, and ${\rm Shape}$. The ${\rm Activity}$ such as ``Drink wine while watching television3'' links to ${\rm Event}$ nodes such as ``event0'' and ``event1'' which indicate events happening in sequence. Each ${\rm Event}$ node connects to the main ${\rm Object}$ node (\eg wine465) and the ${\rm Action}$ node (\eg grab) associated with the event. Additionally, ${\rm Event}$ nodes are connected to ${\rm Situation}$ nodes that describe the state of each object (\ie a ${\rm State}$ node) in the environment before and after the event using the ${\rm StateVal}$ nodes (\eg ON, OFF, CLEAN and DIRTY). Meanwhile, ${\rm Shape}$ nodes represent the 3D coordinates of agents and object states and are linked from ${\rm State}$ nodes by bbox relations and to each other through spatial relations (\eg \textit{inside}, \textit{on}, and \textit{close}).

Since the average number of triplets for a single activity in VirtualHome2KG is over 29,237, we selectively curated and combined only the most relevant triplets within activity KGs as a scenario KG to suit the purpose of our dataset for the task. Instead of including all objects present in the environment, we focused on the main and target objects directly involved in the events of each activity. This approach aligns with our goal of constructing dialogues centered around activities, as most dialogue turns typically pertain to the main and target objects on which actions are performed. Additionally, to enhance the temporal coherence of the scenario, we introduced additional triplets with the \textbf{nextActivity} relationship to link adjacent activities in chronological order.

\subsubsection{Preparation of Scenario Video Summaries}
\begin{table*}[t]
\fontsize{8}{10}\selectfont
    \centering
    \begin{tabular}{c|c|c|c|c|c|c|c|c}
    \hline
    Dataset & \#Videos & \#Dialogues & \#QA Pairs & Video Source & \makecell{Avg. Video \\ Length} & \makecell{Avg. Question \\ Length} & \makecell{Avg. Answer \\ Length} & KG \\ \hline
       VisDial & 120k (images) & 120k & 1.2M & - & - & 5.1 & 8.2 & \xmark \\
       Twitch-FIFA & 49 & 15,083 & 15,083 & Soccer match & 30 secs & 68 & 6.3 & \xmark \\
       AVSD  & 11,816 & 11,816 & 118,160 & Crowdsourced & 30 secs & 7.9 & 9.4 & \xmark \\
       \DATASET  & 1,000 & 3,000 & 30,095 & VirtualHome & 248 secs & 7.8 & 10.2 &\cmark \\ \hline
    \end{tabular}
    \caption{Comparison of our dataset with the existing datasets.}
    \label{tab:dataset-stat}
\end{table*}
We generated video summaries for each scenario by translating triplets from the scenario KGs using a \textbf{template}-based approach. The examples are shown in Table \ref{tab:kg2text-template}.
After generating the linearized KG-to-text summaries, we further \textbf{refined} the text using a commercial LLM, specifically GPT-4o-mini, with the prompt: “Please summarize the following text without adding any extra information: \{text\},” where \{text\} is the placeholder for the linearized summary. This additional step is necessary to remove redundant sentences that might arise from triplets where object states remain unchanged throughout the event sequence. For example, the sentence ``the stove is inside the kitchen.'' might appear repeatedly for multiple events, so this step helps eliminate such redundancies.
After the refinement, we checked the quality of a few summaries whether they contained necessary information and were free from fabricated details.

\subsection{Dataset Analysis}
Table \ref{tab:dataset-stat} compares the existing datasets for the visual dialogue task with ours.
While our dataset includes a smaller number of videos and dialogues, it presents target videos with a longer average duration and represents a diverse combination of activities. Additionally, unlike existing datasets, our dataset includes KGs that provide structured, event-centric information linking activities, events, and other relevant details. These KGs are a valuable resource for developing and evaluating video-grounded dialogue systems.

\subsubsection{Activities for Scenarios}  



\begin{figure}[t]
     \centering
     \begin{subfigure}[b]{0.45\textwidth}
         \centering
         \hfill\includegraphics[width=0.9\textwidth]{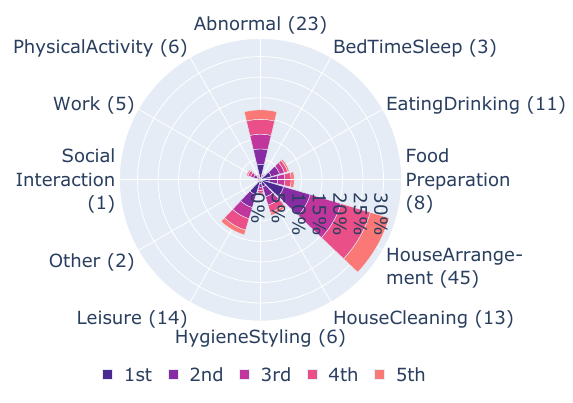}
         \caption{The occurrence of activities by categories based on their placements in scenarios. The number in parentheses indicates the total count of activities within that category.}
         \label{fig:act-cat}
     \end{subfigure} \\ \hfill
     \begin{subfigure}[b]{0.45\textwidth}
         \centering
         \includegraphics[width=0.62\textwidth]{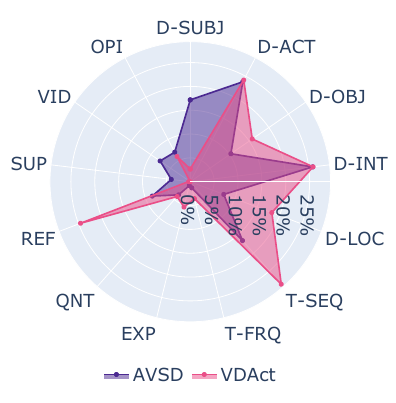}
         \caption{Comparison of the number of question types.}
         \label{fig:q-type-radar}
     \end{subfigure}
     \caption{Statistics on activities as scenarios in the VDAct dataset and percentages of different questions types for sample dialogues in comparison with the AVSD dataset.}
\end{figure}

We analyzed the diversity of activity combinations within scenarios by examining the distribution of activities across categories, as shown in Figure \ref{fig:act-cat}. The analysis indicates a balanced distribution of activities among $1^{\rm st}$ to $4^{\rm th}$ placement within scenarios, categorized by their respective types. Although the HouseArrangement category shows a noticeably higher number of activities in scenarios compared to other categories, this is justified by the fact that it contains nearly twice as many distinct activities (45) as the second-highest category (23).

\subsubsection{Question Types for Dialogues}  



In Figure~\ref{fig:q-type-radar}, we conducted a comprehensive analysis of the question types used in the dataset by categorizing them into 13 pre-defined types based on their distinct characteristics, as detailed in Table \ref{tab:question-type-definition}. To compare our dataset with the closely related AVSD dataset, we randomly selected 60 dialogues from each and labeled each question with one or more of our question types. It is important to note that each question can be assigned multiple types. Our analysis revealed that the AVSD dataset primarily features the question types D-ACT, D-INT, and D-AGT, whereas our dataset highlights T-SEQ, D-INT, REF, and D-ACT as the top types. This difference emphasizes the realistic nature of our dialogues, which frequently reference dialogue history in a natural conversational style. Additionally, our dataset emphasizes discussions about activities involving actions and interactions between actions and objects, including temporal inferences. Particularly, it contains a significantly higher number of questions classified as D-OBJ, D-LOC, T-SEQ, EXP, and REF compared to the AVSD dataset. 
Meanwhile, as opposed to our dataset, the AVSD dataset includes a significantly higher number of questions related to D-AGT and VID types. This is because the AVSD dataset employs crowdsourced videos with varying agents, leading to a greater focus on questions concerning agent states such as ages, appearances, and emotions and the video attributes such as audio and language spoken.

\subsubsection{Scenario KGs}  
\begin{table}[t]
    \centering
    \fontsize{8}{10}\selectfont
    \begin{tabular}{c|c|c}
    \hline
         & Activity & Scenario \\ \hline
         Events & 10.2 & 40.26 \\
         Situations & 11.2 & 44.56 \\
         States & 14.92 & 60.36 \\
         StateVals & 10.84 & 44.04 \\
         Shapes & 86.56 & 360.17 \\
         Main Objects & 4.08 & 13.56 \\
         Target Objects & 0.78 & 2.57 \\
         Triplets & 320.78 & 1,317.51 \\ \hline
    \end{tabular}
    \caption{Comparison of Scenario and Activity KGs by the average number of different components.}
    \label{tab:kg-stats}
    \vspace{-1em}
\end{table}
Table~\ref{tab:kg-stats} shows the statistics of our scenario KGs after merging the activity KGs. From this table, it suggests that each scenario KG contains more number of ${\rm Events}$, ${\rm Situations}$, ${\rm States}$, and ${\rm Objects}$ nodes than the activity KGs. This increase highlights the greater complexity and detail of our dataset in providing compound activities with multiple related events for the dialogue task.

\section{\METRIC{} Metric}
We introduce a new evaluation metric for the video-grounded dialogue task to overcome the shortcomings of the existing metrics. Particularly due to the lack of sufficient context, the existing metrics failed to verify the content in generated responses as they are only presented with the user utterances (\ie questions) and references. Typically, in the context of dialogue, different responses can be considered accurate, even if they do not precisely match the references, as long as they align with the video context and dialogue history. In addition, in some cases, responses may include extra information that needs to be assessed for relevance.

Thus, for our proposed metric, we extended the existing LLM-based metric which demonstrates a high correlation with humans for the VQA task, LAVE \cite{manas2024improving}. LAVE uses rationales and a scaled score rating of 1-3 to assess the generated answer (\ie response) given the question and the reference answer. Our new metric introduces two key improvements over LAVE. First, instead of evaluating the generated response using the individual turn-based context, we include the entire dialogue history, which provides context from the previous question-answer pairs with evaluated scores and rationale.  
Second, we add a video summary from scenario KGs that offers a high-level overview of the video content for evaluation. 
Our approach to incorporating session-based dialogue context is based on the idea that this additional context provides details and information that are difficult to capture with just the question and reference alone. By including the full dialogue history and video content summary, LLMs can use this richer context to accurately assess the correctness of the current response.
\begin{figure}[t]
    \centering
    \includegraphics[width=\linewidth]{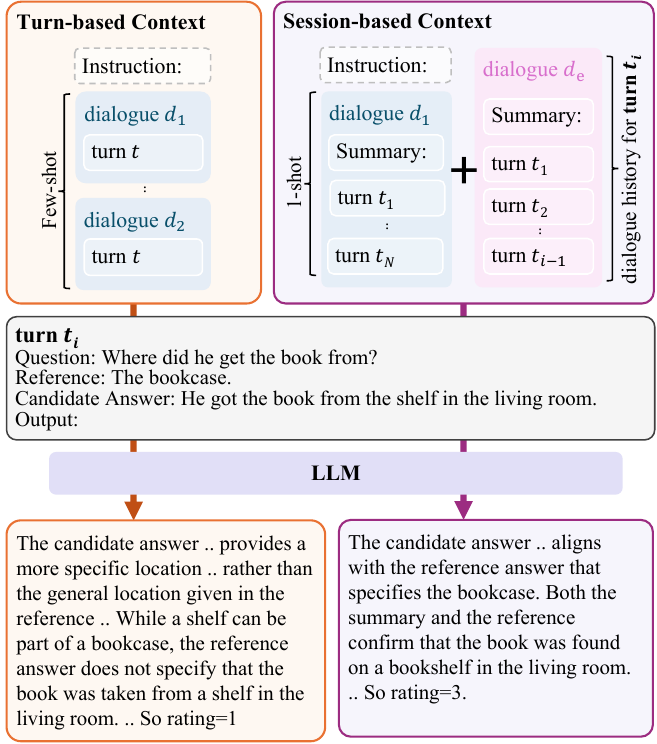}
    \caption{Comparison of turn-based and session-based contexts for evaluation metrics. The input prompt includes turn- or session-based context with the information of turn $t_i$.}
    \label{fig:metric-comparison}
\end{figure}
Figure \ref{fig:metric-comparison} illustrates how our enhanced metric compares to the existing one.

\section{Experiments}
\begin{table*}[t]
\fontsize{8}{10}\selectfont
\centering
\begin{tabular}{cc|cccccccc}
\hline
 & & BLEU & ROUGE & METEOR & SPICE & LLM-Acc & LLM-Rel & LAVE & VDEval \\ \hline
\multicolumn{1}{c|}{\multirow{4}{*}{Frozen}} & Video-LLaVA (8) & 6.56 & 26.68 & 29.24 & 25.15 & 19.74 & 31.94 & 30.14 & 29.01\\
\multicolumn{1}{c|}{} & Video-ChatGPT (100) & 4.38 & 24.09 & 24.88 & 22.87 & 18.99 & 30.89 & 29.48 & 29.56\\
\multicolumn{1}{c|}{} & VideoLLaMA2 (8) & \textbf{6.96} & \textbf{28.86} & \textbf{32.20} & \textbf{27.33} & \textbf{20.99} & \textbf{33.63} & \textbf{32.20} & \textbf{32.89} \\
\multicolumn{1}{c|}{} & VideoLLaMA2 (16) & 6.80 & 27.73 & 30.55 & 26.17 & 19.67 & 32.02 & 31.17 & 31.07\\ \hline
\multicolumn{1}{c|}{\multirow{4}{*}{LoRA finetuned}} & Video-LLaVA (8) & 11.05 & \textbf{\underline{37.16}} & \textbf{\underline{42.08}} & 34.59 & 30.02 & 43.13 & 41.06 & 42.19 \\
\multicolumn{1}{c|}{} & Video-ChatGPT (100) & 8.64 & 33.65 & 38.87 & 31.86 & 23.92 & 36.53 & 34.79 & 35.62\\
\multicolumn{1}{c|}{} & VideoLLaMA2 (8) & \textbf{\underline{11.80}} & 36.63 & 40.87 & \textbf{\underline{35.30}} & \textbf{\underline{30.92}} & \textbf{\underline{44.14}} & \textbf{\underline{41.84}} & \textbf{\underline{43.13}}\\
\multicolumn{1}{c|}{} & VideoLLaMA2 (16) & 10.68 & 35.56 & 40.45 & 34.41 & 30.06 & 42.91 & 41.03 & 42.39\\ \hline
\multicolumn{1}{c|}{\multirow{4}{*}{Proprietary}} & GPT-4o (8) & 5.95 & 26.93 & 27.55 & 25.02 & 27.32 & 40.07 & 38.46 & 37.30 \\
\multicolumn{1}{c|}{} & GPT-4o (16) & \textbf{6.37} & \textbf{27.60} & \textbf{28.14} & \textbf{25.58} & \textbf{30.39} & \textbf{43.25} & \textbf{41.38} & \textbf{41.26} \\
\multicolumn{1}{c|}{} & Gemini-1.5-pro-002 (8) & 1.92 & 21.60 & 19.28 & 21.73 & 21.97 & 36.85 & 37.71 & 34.41 \\
\multicolumn{1}{c|}{} & Gemini-1.5-pro-002 (16) & 1.80 & 21.86 & 19.02 & 21.97 & 24.03 & 39.39 & 40.08 & 37.77 \\ \hline
\end{tabular}
\caption{Performances of baselines across various metrics in terms of dialogue generation. The numbers in parentheses represent the number of video frames used by each baseline. The scores of LLM-Rel, LAVE, and VDEval have been scaled to the 0-1 range and are reported in percentages to ensure consistency with the other metrics. \textbf{Bold} scores represent the highest scores within each baseline group, while \underline{underlined} scores denote the highest scores across all baselines.}
\label{tab:result}
\end{table*}

\begin{table}[t]
\fontsize{8}{10}\selectfont
    \centering
    \begin{tabular}{llccc}
    \hline
    Metric type & & VL2 frozen & VL2 finetuned \\ \hline
    \multirow{4}{*}{\textbf{Classic}} & BLEU & 26.13 & 34.27\\
    & ROUGE & 32.65 & 37.39\\
    & METEOR & 31.35 & 42.73\\
    & SPICE & 30.02 & 38.94 \\\hline
    \multirow{2}{*}{\textbf{Similarity-based}} & BERTScore & 22.43 & 29.59\\
    & BARTScore & 22.29 & 38.45\\\hline
    \multirow{2}{*}{\textbf{LLM-based}} & LAVE & \underline{66.94} & 69.35 \\
    & LLM-Rel & 62.46 & \underline{70.08} \\[-0.9em]\\ \Xhline{3\arrayrulewidth}\\[-0.9em]
    \multicolumn{4}{l}{\textbf{\METRIC{}}~(Ours)} \\
    \multicolumn{1}{c}{Context} & KG summary & \\\hline
    & \multicolumn{1}{c}{\xmark} & 66.85 & 70.09 \\
    \multicolumn{1}{c}{Turn-based}& \multicolumn{1}{c}{Template} & 70.37 & 70.71 \\
    & \multicolumn{1}{c}{Refined} & 70.68 & 71.75\\\hline
    & \multicolumn{1}{c}{\xmark} & 68.25 & 72.59 \\
    \multicolumn{1}{c}{Session-based}& \multicolumn{1}{c}{Template} & 69.12 & 71.63\\
    & \multicolumn{1}{c}{Refined} & \textbf{72.30}${}^*$ & \textbf{73.62}${}^*$\\
    \hline
    \end{tabular}
    \caption{Kendall's rank correlation coefficient between various evaluation metrics and human judgments. VL2 stands for VideoLLaMA2. Bold indicates the highest correlation among all metrics, while underline indicates the highest correlation among existing metrics. * indicates statistically significant improvements in correlation from the underlined scores at the significance level of $p<0.05$.}
    \label{tab:correlation}
\end{table}

\subsection{Experimental Settings}

\paragraph{Data Splits}
Our dataset comprises 3,000 dialogues created from 1,000 scenarios, with each scenario created by three pairs of annotators. To prevent the occurrence of dialogues based on identical scenarios across training and validation/test sets, the dataset is split at the scenario level. We set the fraction of the train, test, and validation as 80\% (2,400 dialogues), 15\% (450), and 5\% (150), respectively.

\paragraph{Baselines}
We adopted state-of-the-art large-scale vision-language foundation models, including both open models such as Video-LLaVA~\cite{lin2023videollava}, Video-ChatGPT~\cite{maaz2024videochatgpt}, and VideoLLaMA2~\cite{damonlpsg2024videollama2}, as well as proprietary models GPT-4o and Gemini-1.5-pro, as baseline models for our dataset. We evaluated open models both with frozen pretrained parameters and with LoRA fine-tuned parameters on the train set of \DATASET{}. We followed the same video frame sampling method used during its pretraining, extracting a fixed number of frames at regular intervals.
For fine-tuning with LoRA and model inferences, we report the parameter settings on the supplementary material.


\paragraph{Baseline Metrics}
We adopted widely used metrics for evaluating generated text: BLEU~\cite{papineni2002bleu}, ROUGE~\cite{lin2004rouge}, METEOR~\cite{banerjee2005meteor}. 
We also included the standard image captioning metric SPICE~\cite{anderson2016spice}, and similarity-based metrics: BERTScore~\cite{zhang2020bertscore}, and BARTScore~\cite{weizhe2021bartscore}.

Additionally, we adopted LLM-based evaluation metrics LLM-Acc/Rel~\cite{maaz2024videochatgpt} and LAVE~\cite{manas2024improving}, specifically used for the VQA task.
LLM-Acc assesses the binary correctness of generated text by feeding QA pairs and predicted answers to LLMs. LLM-Rel rates text quality on a 0-5 scale, with higher scores indicating better quality. LAVE scores 1-3 and employs a few-shot examples as an instruction prompt.

It is noted that the only difference between \METRIC{} with turn-based context without KG summaries and LAVE is a selection of few-shot examples. \METRIC{} uses a 1-shot dialogue example drawn from the \DATASET{} dataset, while LAVE employs few-shot examples provided by the original paper for the general VQA task.
We selected GPT-4o-mini for the LLM-based evaluation metrics LAVE, LLM-Acc/Rel, and \METRIC{}.


\paragraph{Correlation with Human Judgment}
To examine the validity of existing evaluation metrics and our \METRIC{}, we prepared human evaluation judgments and calculated the correlation coefficient between human scores and scores by automatic evaluation metrics.
For human evaluation, we randomly selected 20\% of the test set, comprising 90 dialogues with approximately 900 turns. We defined scores from 1 to 3 as 1: incorrect, 2: partially correct, 3: correct. 
Three independent annotators were employed, with each annotator scoring all 90 dialogues individually. We instructed annotators to comprehensively score the quality of answers by comparing them with gold references and checking the content of the video and dialogue history.
To determine the final score for each sample, we took a majority vote among the scores from three annotators. In cases where all three annotators gave different scores, we excluded those samples from the correlation calculation. Following previous evaluation metrics~\cite{anderson2016spice, wada2024polos}, we adopted the Kendall-B rank correlation coefficient as a correlation metric with human evaluation.
To assess the statistical significance of the improvement in correlation achieved by \METRIC{} compared to existing metrics, we employed a permutation test~\cite{Good2000} with 10,000 iterations.



\subsection{Results and Discussion}
\subsubsection{Baseline Performances on \DATASET{}}
Table~\ref{tab:result} shows a performance comparison of various baseline models on \DATASET{} test set. Notably, all baseline models exhibited performance below 50\% on \METRIC{} metric, indicating that the quality of the generated dialogue answers does not even exceed a score of 2, which means ``partially correct''. These results highlight the difficulty of the \DATASET{} task and underline the need for further advancements in video-grounded dialogue systems for event-driven activities.
When the model parameters are frozen, VideoLLaMA2 8-frame model consistently outperformed other baselines across all evaluation metrics. When model parameters are fine-tuned, VideoLLaVa 8-frame model achieves the highest performance on ROUGE and METEOR metrics, while the VideoLLaMA2 8-frame model showed the best performance on the remaining metrics.
In addition, increasing the number of input video frames for VideoLLaMA2 from 8 to 16 resulted in a performance decrease across all evaluation metrics. This result aligns with a consistent trend reported in the VideoLLaMA2 paper on open-ended VQA benchmarks. These results highlight the current limitations of open VLMs and indicate the need for methodological improvements for better video understanding.

Regarding proprietary models, the GPT-4o 16-frame model achieved comparable performance to the fine-tuned VideoLLaMA2 with both 8 and 16 frames, across all LLM metrics except for VDEval (i.e., LLM-Acc, LLM-Rel, and LAVE). When compared to VideoLLaMA2 with frozen parameters, both GPT-4o variants showed large improvements in LLM metrics. In contrast, for Gemini-1.5-pro, both 8-frame and 16-frame configurations underperformed across every evaluation metric when compared to fine-tuned VideoLLaMA2, although they surpassed the frozen-parameter models in LLM metrics. However, similar to open-source VLMs, the proprietary VLMs still struggled to reach even 50\% on any evaluation metric. This further underscores the difficulty of the VDAct dataset.

\subsubsection{Correlation between Human Judgments}
Table~\ref{tab:correlation} shows the rank correlation coefficients between human assessment and various evaluation metrics, including our proposed \METRIC{}. We can see that LLM-based evaluation metrics showed higher correlations compared to classic metrics and similarity-based metrics, indicating the effectiveness of using LLMs for evaluating dialogue systems.
Regarding \METRIC{}, we first found that incorporating session-based context improved correlation coefficients regardless of the use of KG summaries. Regarding KG summaries, incorporating template-based summaries led to a decrease in correlation coefficients for the fine-tuned VideoLLaMA2 with the session-based context, while refined summaries consistently enhanced correlation coefficients across all experimental settings.
In particular, the variation that uses both the session-based context and refined KG summaries showed the highest correlation with human evaluation, achieving a significant improvement compared to the existing state-of-the-art metrics, substantiating the effectiveness of \METRIC{} in more accurately reflecting human judgments in video-grounded dialogue evaluation.

\subsubsection{Performances on Different Question Types}
We evaluated the performance of VideoLLaMA2, the best fine-tuned baseline on \DATASET{}, across various question types.
Our analysis revealed that the baseline performed less effectively on T-SEQ, T-FRQ, and EXP question types, with average \METRIC{} scores of 1.49 (24.5\%), 1.56 (28\%), and 1.62 (31\%), respectively. 
In contrast, the average scores for questions not falling into these types were higher at 1.7 (35\%), 1.66 (33\%), and 1.66 (33\%). 
For QNT questions, the baseline showed better performance, with an average score of 1.78 (39\%) compared to 1.65 (32.5\%) for questions that do not belong to QNT. 
This better performance is likely because many QNT questions involve determining a small number of items (1 or 2) as the correct answer.
Thus, this simplifies the system's task for the QNT question type.
Additionally, the baseline performed better on binary questions as it achieved an average score of 1.78 (39\%), compared to 1.59 (29.5\%) for other question types. 

\subsubsection{The Effect of Question and Answer Length}
We examined how the length of questions and answers affects the performance of the best fine-tuned baseline.
Our findings show that the average word counts of reference answers where their generated responses achieved VDEval scores of 1 (0\%), 2 (50\%), and 3 (100\%) are 11.51, 12.83, and 7.52 words, respectively. This indicates that the model struggles to provide accurate responses for more complex questions as they require more elaborate answers.
To further support the above claim, the third quartile regarding the word counts of answers for questions with a score of 3 (100\%) is 10 words, while for others with scores of 1 (0\%) and 2 (50\%), the third quartiles are 15 and 16 words, respectively. This suggests that the model’s accuracy declines as the length of the reference answers increases.

\section{Conclusion}
This paper presents VDAct, a novel dataset for video-grounded dialogue on event-driven activities, and VDEval, a specialized evaluation metric for this task. 
VDAct features longer and more complex video sequences depicting diverse activity scenarios that demand advanced contextual understanding. Experimental results showed that \DATASET{} task is challenging and includes several question types that are difficult to answer.
Additionally, VDEval, which incorporates dialogue session history and video content summaries from KGs for evaluating dialogue responses, demonstrates a significantly higher correlation with human assessments compared to existing metrics. 
Future work can explore a model architecture that utilizes KGs for video-grounded dialogue.

\section{Acknowledgment}
This paper is based on results obtained from: (1) a project, Programs for Bridging the gap between R\&D and the IDeal society (society 5.0) and Generating Economic and social value (BRIDGE)/Practical Global Research in the AI × Robotics Services, implemented by the Cabinet Office, Government of Japan, and (2) a project, JPNP20006, commissioned by the New Energy and Industrial Technology Development Organization (NEDO).

In addition, we would like to thank Susan Holm, our knowledge engineer, for her insightful feedback during the data collection process.

\bibliography{aaai25}

\clearpage

\appendix
\section{Technical Appendix}
\subsection{Additional Dataset Analysis}
\subsubsection{Sample Events, Activities, and Activity Types for a Scenario}
We selected a sample scenario from VDAct and detailed it with specific activities and their corresponding sequence of events, as illustrated in Table \ref{tab:activity-type-event}.
\begin{table}[H]
\fontsize{8}{9}\selectfont
    \centering
    \begin{tabular}{p{0.3\linewidth}l}
    \hline
    \multirow{7}{*}{\makecell[l]{Activity \#1: \\ Pick up dirty fork1 \\ (HouseArrangement)}} & [WALK] $\langle$fork$\rangle$ (218) \\
    & [GRAB] $\langle$fork$\rangle$ (218) \\
    & [WALK] $\langle$fork$\rangle$ (222) \\
    & [GRAB] $\langle$fork$\rangle$ (222) \\
    & [WALK] $\langle$sink$\rangle$ (197) \\
    & [PUTBACK] $\langle$fork$\rangle$ (218) $\langle$sink$\rangle$ (197) \\
    & [PUTBACK] $\langle$fork$\rangle$ (222) $\langle$sink$\rangle$ (197) \\ \hline

    \multirow{9}{*}{\makecell[l]{Activity \#2: \\ Pick up dirty waterglass1 \\ (HouseArrangement)}} & [WALK] $\langle$waterglass$\rangle$ (216) \\
    & [GRAB] $\langle$waterglass$\rangle$ (216) \\
    & [WALK] $\langle$waterglass$\rangle$ (220) \\
    & [GRAB] $\langle$waterglass$\rangle$ (220) \\
    & [WALK] $\langle$sink$\rangle$ (197) \\
    & [POUR] $\langle$waterglass$\rangle$ (216) $\langle$sink$\rangle$ (197) \\
    & [POUR] $\langle$waterglass$\rangle$ (220) $\langle$sink$\rangle$ (197) \\
    & [PUTBACK] $\langle$waterglass$\rangle$ (216) $\langle$sink$\rangle$ (197) \\
    & [PUTBACK] $\langle$waterglass$\rangle$ (220) $\langle$sink$\rangle$ (197) \\ \hline
    
    \multirow{12}{*}{\makecell[l]{Activity \#3: \\ Clean sink4 \\ (HouseCleaning)}} & [WALK] $\langle$sink$\rangle$ (1) \\
    & [WALK] $\langle$sponge$\rangle$ (1) \\
    & [GRAB] $\langle$sponge$\rangle$ (1) \\
    & [WALK] $\langle$sink$\rangle$ (1) \\
    & [TURNTO] $\langle$sink$\rangle$ (1) \\
    & [WIPE] $\langle$sponge$\rangle$ (1) $\langle$sink$\rangle$ (1) \\
    & [WALK] $\langle$faucet$\rangle$ (1) \\
    & [SWITCHON] $\langle$faucet$\rangle$ (1) \\
    & [SWITCHOFF] $\langle$faucet$\rangle$ (1) \\
    & [WALK] $\langle$sink$\rangle$ (1) \\
    & [TURNTO] $\langle$sink$\rangle$ (1) \\
    & [WIPE] $\langle$sponge$\rangle$ (1) $\langle$sink$\rangle$ (1) \\ \hline

    \multirow{19}{*}{\makecell[l]{Activity \#4: \\ Put groceries in fridge23 \\ (HouseArrangement)}} & [WALK] $\langle$milk$\rangle$ (1) \\
    & [GRAB] $\langle$milk$\rangle$ (1) \\
    & [PUTOBJBACK] $\langle$milk$\rangle$ (1) \\
    & [WALK] $\langle$bedroom$\rangle$ (1) \\
    & [WALK] $\langle$bathroom$\rangle$ (1) \\
    & [WALK] $\langle$towel$\rangle$ (1) \\
    & [GRAB] $\langle$towel$\rangle$ (1) \\
    & [WALK] $\langle$bedroom$\rangle$ (1) \\
    & [WALK] $\langle$kitchen$\rangle$ (1) \\
    & [WALK] $\langle$milk$\rangle$ (1) \\
    & [GRAB] $\langle$milk$\rangle$ (1) \\
    & [SCRUB] $\langle$milk$\rangle$ (1) $\langle$towel$\rangle$ (1) \\
    & [WALK] $\langle$kitchentable$\rangle$ (1) \\
    & [PUTBACK] $\langle$towel$\rangle$ (1) $\langle$kitchentable$\rangle$ (1) \\
    & [WALK] $\langle$fridge$\rangle$ (1) \\
    & [OPEN] $\langle$fridge$\rangle$ (1) \\
    & [TURNTO] $\langle$fridge$\rangle$ (1) \\
    & [PUTIN] $\langle$milk$\rangle$ (1) $\langle$fridge$\rangle$ (1) \\
    & [CLOSE] $\langle$fridge$\rangle$ (1) \\ \hline

    \multirow{9}{*}{\makecell[l]{Activity \#5: \\ Fall while preparing \\meal1 (Abnormal)}} & [WALK] $\langle$milk$\rangle$ (1) \\
    & [GRAB] $\langle$milk$\rangle$ (1) \\
    & [WALK] $\langle$cupcake$\rangle$ (1) \\
    & [GRAB] $\langle$cupcake$\rangle$ (1) \\
    & [WALK] $\langle$bedroom$\rangle$ (1) \\
    & [WALK] $\langle$livingroom$\rangle$ (1) \\
    & [WALK] $\langle$coffeetable$\rangle$ (1) \\
    & [TURNTO] $\langle$remotecontrol$\rangle$ (1) \\
    & [FALL] \\ \hline
    
    \end{tabular}
    \caption{A list of activities for a scenario with their associated sequence of events. Each event consists of an action (\eg [POUR]), main object (\eg $\langle$waterglass$\rangle$), and target object (\eg $\langle$sink$\rangle$). Each object is presented with its ID.}
    \label{tab:activity-type-event}
\end{table}

\subsubsection{Sample Scenarios and Dialogues}
Table \ref{tab:example-dials} and \ref{tab:example-dials2} present scenarios and their corresponding dialogues created by three different annotator pairs. Each scenario represents a different sequence of activities. The dialogues present the different information exchange between each annotator pair, although the scenario and its activities are the same.

\begin{table*}[t]
\fontsize{7}{8}\selectfont
\centering
\begin{tabular}{c|p{0.28\linewidth}|p{0.28\linewidth}|p{0.28\linewidth}}
\hline \hline
\multicolumn{4}{l}{\textbf{Example}\#1, \textbf{Scenario ID}: 00265, \textbf{Activities}: Pick up dirty fork1, Pick up dirty waterglass1, Clean sink4, Put groceries in fridge23, Fall while preparing meal1} \\ \hline
 & \textbf{Dialogue ID}: 0026501 & \textbf{Dialogue ID}: 0026502 & \textbf{Dialogue ID}: 0026503 \\ \hline
\multirow{2}{*}{1} & Was the man carrying anything when he fell? & What room did the man start his activities in? & Where does he begin at today? \\ \cline{2-4} 
 & Yes, he was. & In the kitchen. & He finds himself in the kitchen manhandling some forks. \\ \hline
\multirow{2}{*}{2} & Was he carrying a frying pan? & What is the first thing he did there? & Why is he handling forks? \\ \cline{2-4} 
 & Yes, he had a pan in his right hand. & He picks up two dirty forks. & I guess they are dirty and decides to take them to the sink. \\ \hline
\multirow{2}{*}{3} & Was he moving to or from the stove when he fell? & What does he do with them? & Does he use water to wash them or just sets them there? \\ \cline{2-4} 
 & He was facing the direction of the stove, but it's unclear where he was taking the items he was holding. & He places them in the sink. & He just sets them there and then goes back to the table for some glasses. \\ \hline
\multirow{2}{*}{4} & Was he cooking something on the stove? & What does he do next? & What does he do with the glasses? \\ \cline{2-4} 
 & No, nothing is sitting on the burners. & He places dirty water glasses in the sink too. & He also takes them to the sink and maybe rinses them out but I can't see any running water. \\ \hline
\multirow{2}{*}{5} & How many items does the man put into the fridge? & Does he do anything else when he's at the sink? & Okay, so after this what does he decide to get into? \\ \cline{2-4} 
 & One item. & He also cleans the sink. & Well the glasses and forks vanish and decides to sponge down the sink. \\ \hline
\multirow{2}{*}{6} & Did the man put a non-food item into the fridge? & What does he use to clean it? & Does he get the sponge wet first? \\ \cline{2-4} 
 & No, it is an edible item. & A sponge. & No he dry wipes first and then uses some water and wipes again. \\ \hline
\multirow{2}{*}{7} & Where does he put the dirty water glass? & Does he do anything else in the kitchen after cleaning the sink? & How long did it take him to do this? \\ \cline{2-4} 
 & In the kitchen sink. & He cleans a carton of milk before putting it in the fridge. & Just a second as he got distracted by the cartoon of TruMoo sitting out behind the sink. \\ \hline
\multirow{2}{*}{8} & Did he fill it with water after putting it in the sink? & What else does he put in the fridge? & What's he do with the TruMoo? \\ \cline{2-4} 
 & Nope. & He doesn't put anything else in the fridge. & He goes to get a towel to wipe it down and then he puts in the sink. \\ \hline
\multirow{2}{*}{9} & How many dirty forks does the man pick up? & What does he do next? & So after he finishes that task where does he go? \\ \cline{2-4} 
 & Two dirty forks. & He picks up a frying pan and package of chicken before falling. & He decides he needs to cook something so he grabs a pan. \\ \hline
\multirow{2}{*}{10} & Did he get them from the living room? & How does he fall? & What does he cook in the pan? \\ \cline{2-4} 
 & No, they were in the kitchen. & He falls forward onto his face. & He never makes it that far as he face plants and then lays on the floor forever. \\ \hline \hline
 \multicolumn{4}{l}{} \\
\multicolumn{4}{l}{\textbf{Example}\#2, \textbf{Scenario ID}: 00523, \textbf{Activities}: Drink alcohol1, Pick up dirty dishes1, Relax on sofa1, Fall while standing quietly1, Cook salmon1} \\ \hline
 & \textbf{Dialogue ID}: 0052301 & \textbf{Dialogue ID}: 0052302 & \textbf{Dialogue ID}: 0052303 \\ \hline
\multirow{2}{*}{1} & What room do you see the man in initially? & What activity did the man start by doing? & What is his first action? \\ \cline{2-4} 
 & Initially, the living room, but it looks like he was moving through the living room to the kitchen. & He drank some alcohol first. & Today he is in the kitchen looking for a drink. \\ \hline
\multirow{2}{*}{2} & Does he make it to the kitchen safely? & What kind of alcohol did he drink? & Does he find one? \\ \cline{2-4}
 & Yes, where liquid refreshment awaits him. & Something from a clear bottle of liqueur. & He does yes, his favorite curacao, he takes a sip. \\ \hline
\multirow{2}{*}{3} & What kind of beverage is there for him to enjoy? & Where did he drink it? & What does he do after drinking? \\ \cline{2-4}
 & Some kind of alcohol in a clear bottle. & In the kitchen. & He checks out the plates on the table, they seem dirty. \\ \hline
\multirow{2}{*}{4} & How many total sips does he have of the alcohol? & Was his next activity in the kitchen too? & How does he clean them then? \\ \cline{2-4}
 & Two sips plus a long pull on the bottle. & Yes, it was. & Well he takes them to the sink for a nice soaking. \\ \hline
\multirow{2}{*}{5} & What is next on the man's agenda? & What did he do in the kitchen? & How long does he soak them for? \\ \cline{2-4}
 & He relocates some dirty dishes to the sink. & He picked up two dirty dishes and put them in the sink. & I guess the rest of the day because he's off to the living room now for a rest. \\ \hline
\multirow{2}{*}{6} & Does he remain in the kitchen for any other reason? & What did the man do after that? & Where does he rest there? \\ \cline{2-4}
 & No, he bee lines for the living room after depositing the plates in the sink. & Next, he went into the living room and relaxed on the sofa for a bit. & On the couch. \\ \hline
\multirow{2}{*}{7} & What awaits him in the living room? & What else did he do in the living room? & What does he do after resting? \\\cline{2-4}
 & Relaxation on his sofa. & He grabbed a couple of books off his bookshelf but had a bit of an issue after that. & He gets up and heads over to the book shelf and grabs two books. \\ \hline
\multirow{2}{*}{8} & Does he stay there for the remainder of his day? & What was his issue? & What were the books could you tell? \\ \cline{2-4}
 & No, but he stays for a while longer because he decided to read a couple of books. & He fell over while holding the books. & I couldn't tell because he falls flat on his face as soon as he gets the book from the shelf. \\ \hline
\multirow{4}{*}{9} & Does he sit to read those chosen books? & Did he recover and move on to his last activity? & Is he able to get back up though? \\ \cline{2-4}
 & Don't know because he just all of a sudden falls over while standing there holding the books. & Yes, he went into the kitchen to do some cooking. & Yes, and he quickly heads to the kitchen for some cooking now. \\ \hline
\multirow{2}{*}{10} & Does he recover from his fall? & What was he cooking in the kitchen? & What does he decide to cook? \\ \cline{2-4}
 & Yes, and heads into the kitchen to do some cooking. & A packet of salmon. & He's got his eye on some salmon. He washes it off in the package and takes it to the stove. \\ \hline
 \multirow{2}{*}{11} & What does he cook? & How did he cook it? & -- \\ \cline{2-4}
 & He is cooking some salmon. & On the stove, using a frying pan. & -- \\ \hline
\end{tabular}
\caption{Dialogues created by three separate annotator pairs in VDAct for scenario IDs 00265 and 00523.}
\label{tab:example-dials}
\end{table*}

\begin{table*}[t]
\fontsize{7}{8}\selectfont
\centering
\begin{tabular}{c|p{0.28\linewidth}|p{0.28\linewidth}|p{0.28\linewidth}}
\hline \hline
\multicolumn{4}{l}{\textbf{Example}\#3, \textbf{Scenario ID}: 00733, \textbf{Activities}: Turn off light5, Drink wine1, Put groceries in fridge17, Clean stove1, Place wineglass1} \\ \hline
 & \textbf{Dialogue ID}: 0073301 & \textbf{Dialogue ID}: 0073302 & \textbf{Dialogue ID}: 0073303 \\ \hline
\multirow{2}{*}{1} & What sorts of things is the man busy doing? & Did the man start in the kitchen? & Where’s the man starting his day? \\ \cline{2-4} 
 & Drinking, cleaning, putting things away, turning off lights, the usual. & Yes, he did. & He's starting in the kitchen of course. \\ \hline
\multirow{2}{*}{2} & What room does he turn the light off in? & What did he get up to first? & He got any plans in there? \\ \cline{2-4}
 & The kitchen. & First, he turned off the light to the kitchen and went into the living room/office. & Actually no as he just turns the lights out and leaves. \\ \hline
\multirow{2}{*}{3} & Had he been busy in the kitchen prior to turning the light off? & What did he do in that part of his house? & Where’s he go to next? \\ \cline{2-4}
 & No, he just turned off the light and went to the living room for no apparent reason. & Nothing, he just walked back into the kitchen, where the light was on again, and drank some wine. & He goes to the living room but reverses course immediately and goes back to the kitchen. \\ \hline
\multirow{2}{*}{4} & Is turning off the light the first thing he was seen doing? & How did he drink the wine? & Do you think he forgot something? \\ \cline{2-4}
 & Yes. & He drank it mostly from a wine glass but also a little from the bottle too. & Yes he forgot about wine which he decides to drink a ton of. \\ \hline
\multirow{2}{*}{5} & So, after going into the living room he returns to the kitchen? & How long did he spend drinking? & Did he drink it right in the kitchen? \\ \cline{2-4}
 & Yes, to have a drink of wine. & About 45 seconds, give or take. & Yes and then he goes to grab a big bag of potato chips. \\ \hline
\multirow{2}{*}{6} & Does he pour himself a glass, or drink it out of the bottle like a savage? & What does he do next? & Does he munch on those as well? \\ \cline{2-4}
 & He does both. & Next, he put a bag of potato chips in the fridge. & Again he fakes us all out and just sticks them in the fridge. \\ \hline
\multirow{2}{*}{7} & Are the rest of his activities done in the kitchen? & Where did he get it from originally? & Does he stay in the kitchen after that? \\\cline{2-4}
 & Yes, they are. & A high shelf on the wall of the kitchen. & Yea he decides to do some cleaning now. \\ \hline
\multirow{2}{*}{8} & What are they? & What else does he do in the kitchen? & What's he cleaning with? \\ \cline{2-4}
 & He puts some chips into the fridge, cleans the stove, and sets out a wine glass, in that order. & He cleans the stove and places a wine glass. & He grabs a sponge and starts working on the stove. \\ \hline
\multirow{4}{*}{9} & Where did the chips come from? & How does he clean the stove? & Do you think he did a good job? \\ \cline{2-4}
 & A floating shelf near the range. & Nah but he's already fixated on another wineglass so guess it doesn't matter. & By wiping it with a sponge. \\ \hline
\multirow{2}{*}{10} & What does he clean the stove with? & Where does he place the wineglass? & What's wrong with the wineglass? \\ \cline{2-4}
 & A white sponge. & From the TV stand onto the kitchen table. & Nothing I don't think he just decides to move it to the table. \\ \hline \hline
 \multicolumn{4}{l}{} \\
\multicolumn{4}{l}{\textbf{Example}\#4, \textbf{Scenario ID}: 00912, \textbf{Activities}: Fall while standing and turning1, Use phone3, Cook fried bread4, Place waterglass1} \\ \hline
 & \textbf{Dialogue ID}: 0091201 & \textbf{Dialogue ID}: 0091202 & \textbf{Dialogue ID}: 0091203 \\ \hline
\multirow{2}{*}{1} & What is the man doing when first seen? & What was the man's first activity? & What's the first thing the home skillet does this fine day? \\ \cline{2-4} 
 & He is seen going into the living room, picking up the TV remote, and as he is turning around tripping or slipping and falling down. & He had a fall in the living room. & He begins in the kitchen. \\ \hline
\multirow{2}{*}{2} & Okay, does he resume doing something in the living room after the fall? & How did that happen? & What's he got going on there? \\ \cline{2-4} 
 & Yes, he picks up his cell phone. & Not entirely sure, he was just walking when he fell forward. & He isn't there for long as he starts wandering around the house. \\ \hline
\multirow{2}{*}{3} & Does he sit down on the sofa to use it? & Was he able to move on to his next activity? & How many rooms does he wonder to? \\ \cline{2-4} 
 & Yes, but only after returning from using it in the bedroom. & Yes, he was. & About two. He finally arrives in the living room to which he goes over and turns on nightmare bunny. \\ \hline
\multirow{2}{*}{4} & Where did the man get the phone? & What was his next activity? & Why would he do that again? \\ \cline{2-4} 
 & From the coffee table in the living room. & He used his phone a couple of times. & Bored. After that he grabs the remote and attempts to sit down with it. \\ \hline
\multirow{2}{*}{5} & Where does he go after using the phone on the sofa? & Where did he pick up his phone from? & Does he not succeed? \\ \cline{2-4} 
 & To the kitchen. & It was next to the TV in the living room. & Nope, he falls down face first. \\ \hline
\multirow{2}{*}{6} & What does his visit to the kitchen consist of? & Where did he use it? & Did he hit his head on anything? \\ \cline{2-4} 
 & He does some cooking and some relocation of two water glasses. & In the bedroom and in the living room. & He seems fine, now he grabs his cellphone and takes it to his bedroom where he plays with it there on the bed. \\ \hline
\multirow{2}{*}{7} & Does he cook or handle the glasses first? & What did he do after using his phone? & What do you think he is playing? \\ \cline{2-4} 
 & He does the cooking first. & He went to the kitchen to do some cooking. & Tetris as per usual. He only spends a couple of seconds with it on the bed and after that he takes it back to the couch to play with it there too. \\ \hline
\multirow{2}{*}{8} & What is the man cooking? & What did he cook? & Does he continue playing Tetris? \\ \cline{2-4} 
 & Some French toast. & Some fried bread with milk. & No. After that he gets real hungry so he goes to the kitchen to make his famous burnt toast and milk on the stove. \\ \hline
\multirow{2}{*}{9} & How many ingredients are used for the dish? & Did he cook it on the stove? & Does he finally enjoy it? \\ \cline{2-4} 
 & Two, toast and milk. & Yes, he did. & Maybe he did but the next thing I know is he's getting two water glasses. \\ \hline
\multirow{2}{*}{10} & From where does the man relocate the water to and from? & What did he do last? & Does he double fist drinking water? \\ \cline{2-4} 
 & From two different spots on the kitchen table to another place on the same table. & He placed some water glasses on the kitchen table. & All he does is pick them up and then set them on the table. \\ \hline
\end{tabular}
\caption{Dialogues created by three separate annotator pairs in VDAct for scenario IDs 00733 and 00912.}
\label{tab:example-dials2}
\end{table*}

\subsubsection{Co-occurrence of Activities as Scenarios}
\begin{figure}[H]
    \centering
    \includegraphics[width=\linewidth]{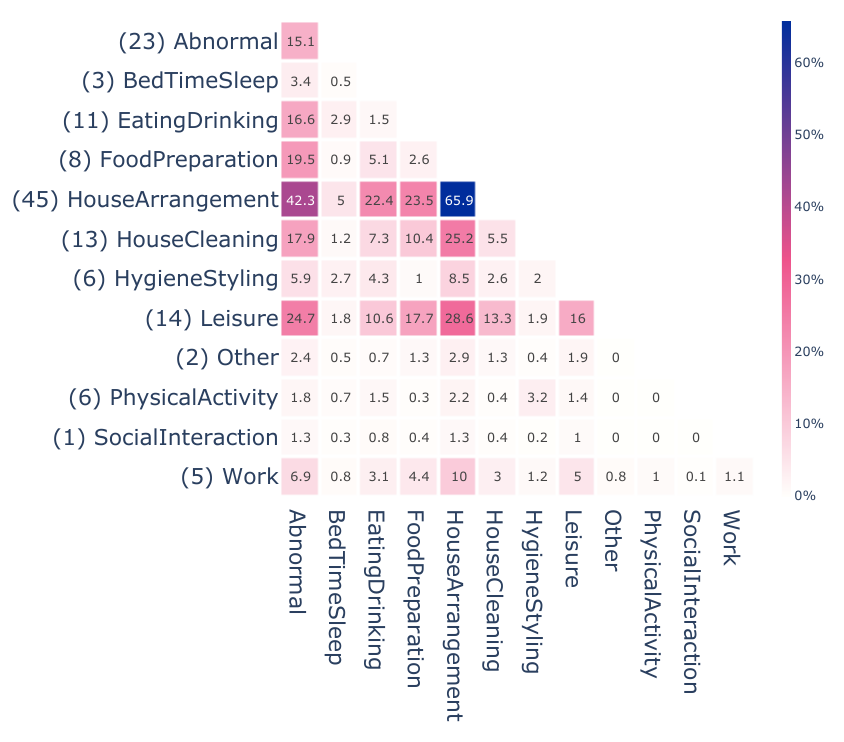}
    \caption{The co-occurrence of activity pairs by their activity types within the scenarios. The number in the parentheses behind each activity type indicates the total number of activities. The number in each grid represents a percentage of an activity-type pair that occurred across scenarios.}
    \label{fig:co-occurrence-activity-types}
\end{figure}

Figure \ref{fig:co-occurrence-activity-types} illustrates the co-occurrences of activity pairs by their activity types regardless of their $i$-th placement within the scenarios. The top three activity-type pairs that appear within the scenarios are HouseArrangement-HouseArrangement (65.9\%), HouseArrangement-Abnormal (42.3\%), and Leisure-HouseArrangement (28.6\%). Meanwhile, due to a small number of activities ($<$ 5) within BedTimeSleep, PhysicalActivity, SocialInteraction, and Other, activities from these types rarely co-occur with other types or with each other.

\subsubsection{Distribution of n-grams for Questions and Answers}
\begin{figure}[H]
    \centering
    \includegraphics[width=\linewidth]{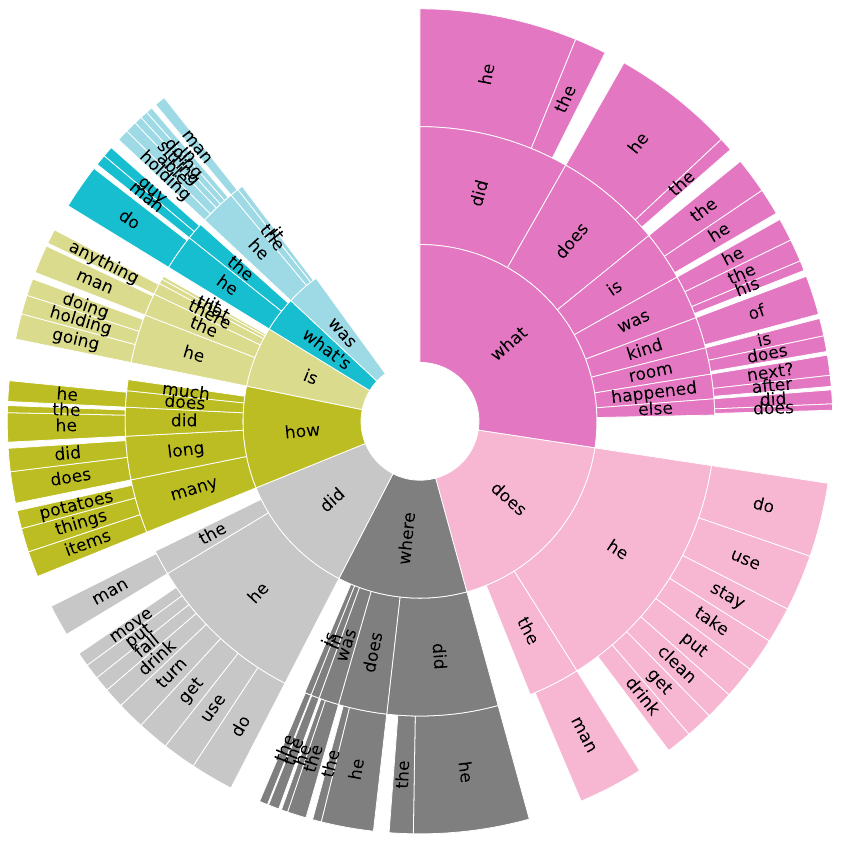}
    \caption{Distribution of \textit{questions} in VDAct based on their first three words. The size of each arc reflects the frequency of each three-word prefix.}
    \label{fig:prefix-count-questions}
\end{figure}

\begin{figure}[H]
    \centering
    \includegraphics[width=\linewidth]{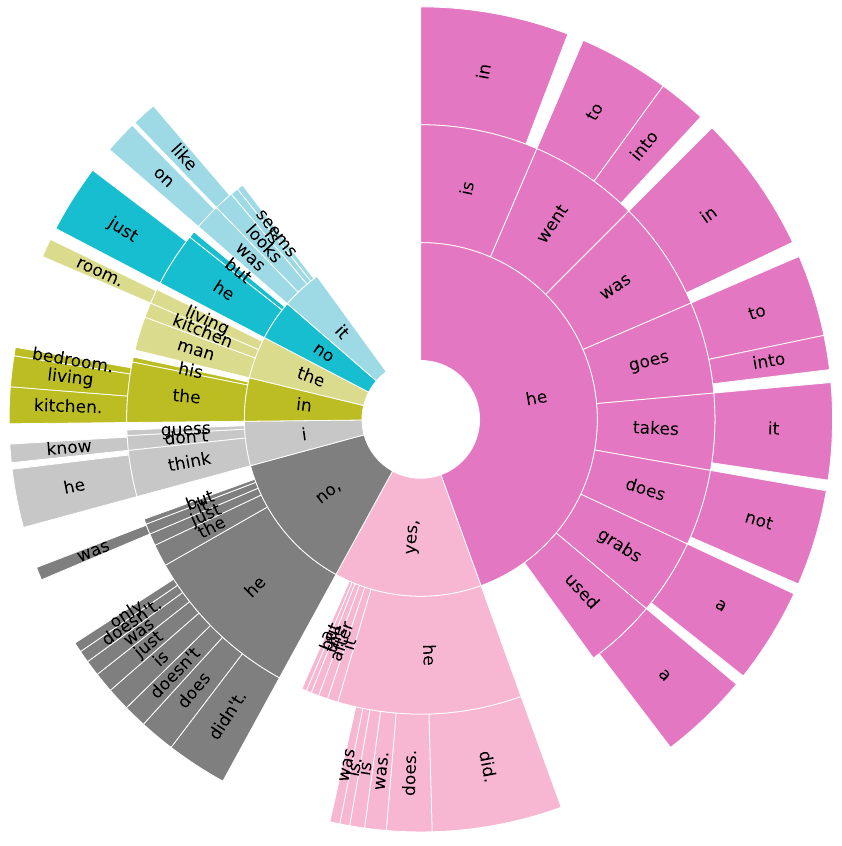}
    \caption{Distribution of \textit{answers} in VDAct based on their first three words. The size of each arc reflects the frequency of each three-word prefix.}
    \label{fig:prefix-count-answers}
\end{figure}

Figure \ref{fig:prefix-count-questions} and \ref{fig:prefix-count-answers} show the distributions of questions and answers by their three-word prefixes. For questions, VDAct includes the top five first words beyond descriptive and binary types. These include ``how'' for various question types such as temporal (\eg ``how long''), quantitative (\eg ``how many/much''), and explanatory (\eg ``how does/did''). For answers, the most common first-word group starts with ``he'', reflecting descriptions of the virtual agent’s movements (\eg ``he is/was in'' and ``he goes/went to/into'') and interactions with objects (\eg ``he used'' and ``he takes''). The second and third most common groups represent binary responses (\eg ``yes, he did'' and ``no, he didn't'').

\subsection{Experimental Details}

\subsubsection{Baseline Models}
This section reviews state-of-the-art Visual Language Models (VLMs) used as baselines on VDAct.
\paragraph{Video-LLaVA} This model aligns images and videos before projection, allowing the LLM to learn from a unified visual representation and comprehend both images and videos simultaneously. The model employs Vicuna-7B v1.5 as the LLM, uses LanguageBind-derived visual encoders (initialized from ViT-L/14), and shared projection layers consisting of two fully connected layers.
\paragraph{Video-ChatGPT} This model leverages the CLIP-L/14 visual encoder to extract both spatial and temporal video features by averaging frame-level features across temporal and spatial dimensions respectively, then feeds the computed spatiotemporal features into a learnable linear layer projecting them into the input space of the Vicuna-7B-v1.1 model which serves as the LLM.
\paragraph{VideoLLaMA2} In this model, the video frames are encoded into features, processed through a spatial-temporal convolution connector, and fed into LLMs. ViT-L/14 is chosen as a vision backbone, and Mistral-7B-Instruct-v0.2 is used as an LLM backbone.

\subsubsection{Training and Inference}
Table~\ref{tab:hyper-params} shows the hyperparameters used when training and inference of VLMs. During training, we adopted 8-bit quantized LoRA fine-tuning, targeting all trainable parameters other than word embeddings and the language modeling head. Additionally, we adopted DeepSpeed ZeRO-3 CPU offloading to manage memory allocation during the fine-tuning.
We fine-tuned the baseline models with a total of 5 epochs.
The best model checkpoint was chosen based on its performance on the SPICE metric that is evaluated using the VDAct development set.
By using this criterion, the best checkpoint for fine-tuned Video-LLaVA, Video-ChatGPT, VideoLLaMA2(8) and VideoLLaMA2(16) are at epoch 5, 4, 4, and 3, respectively.
We used AdamW as the optimizer and eight NVIDIA A100 40GB GPUs for both training and inference. The reported scores are from a single run with a fixed seed.
\begin{table}[H]
    \centering
    \fontsize{8}{9}\selectfont
    \begin{tabular}{lr||lr}
    \hline
    \textbf{Training} & & \textbf{Inference} & \\ \hline
    Learning rate & 2e-05 & Sampling & True \\
    Total mini-batch size & 128 & Temperature & 0.1 \\
    Max sequence length & 2,048 & Top p & 1.0 \\
    LoRA $\alpha$ & 256 & Top k & 0 \\
    LoRA $r$ & 128 & Max new tokens & 256\\
    Random seed & 42 & & \\
    \hline
    \end{tabular}
    \caption{Hyperparameter settings for training and inference on the baseline models.}
    \label{tab:hyper-params}
\end{table}

\subsubsection{Prompts for Evaluation Metrics}
\begin{figure*}[ht]
    \centering
    \includegraphics[width=\linewidth]{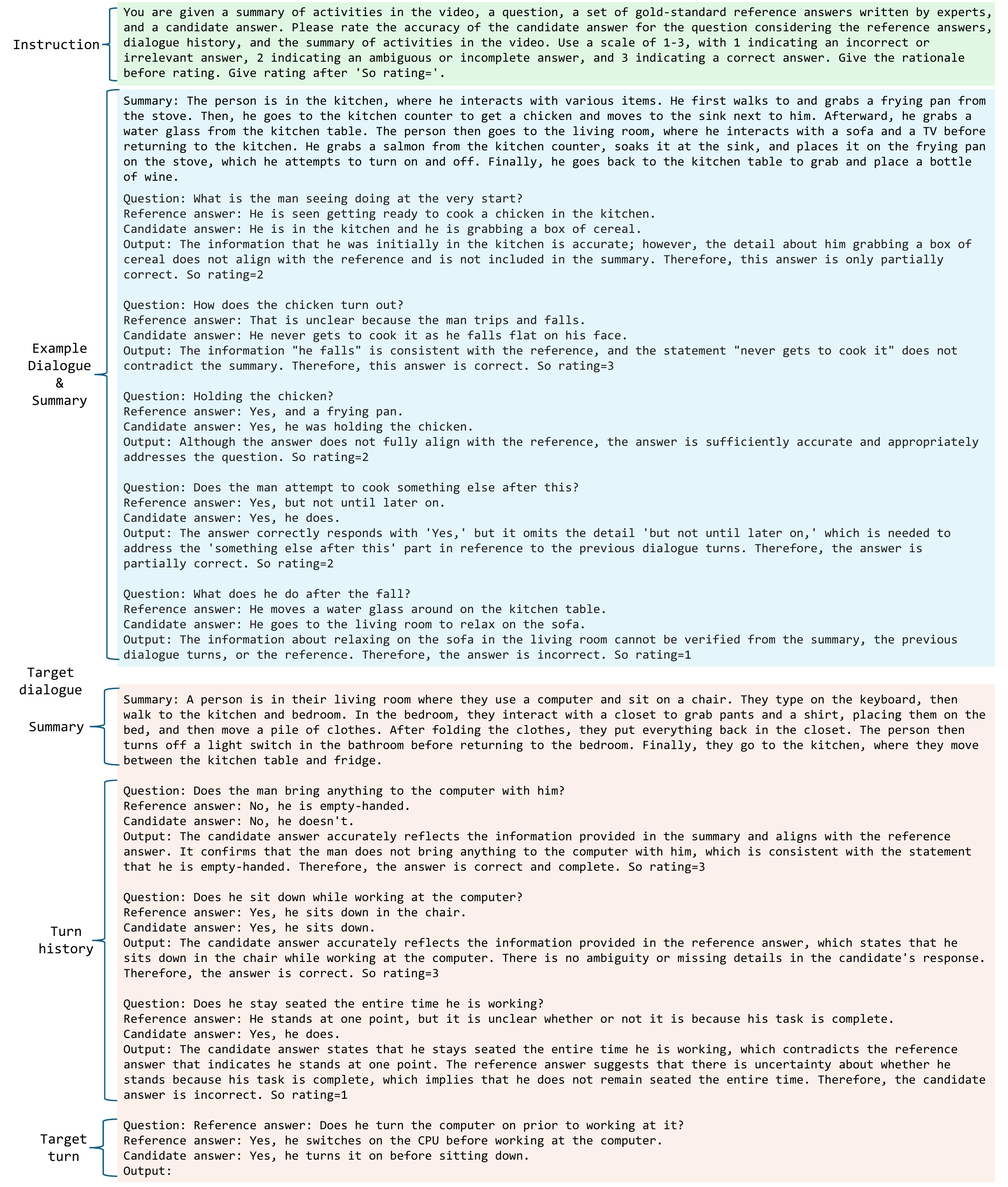}
    \caption{Whole prompt for VDEval. }
    \label{fig:prompt-vdeval}
\end{figure*}

\begin{figure*}[ht]
    \centering
    \includegraphics[width=\linewidth]{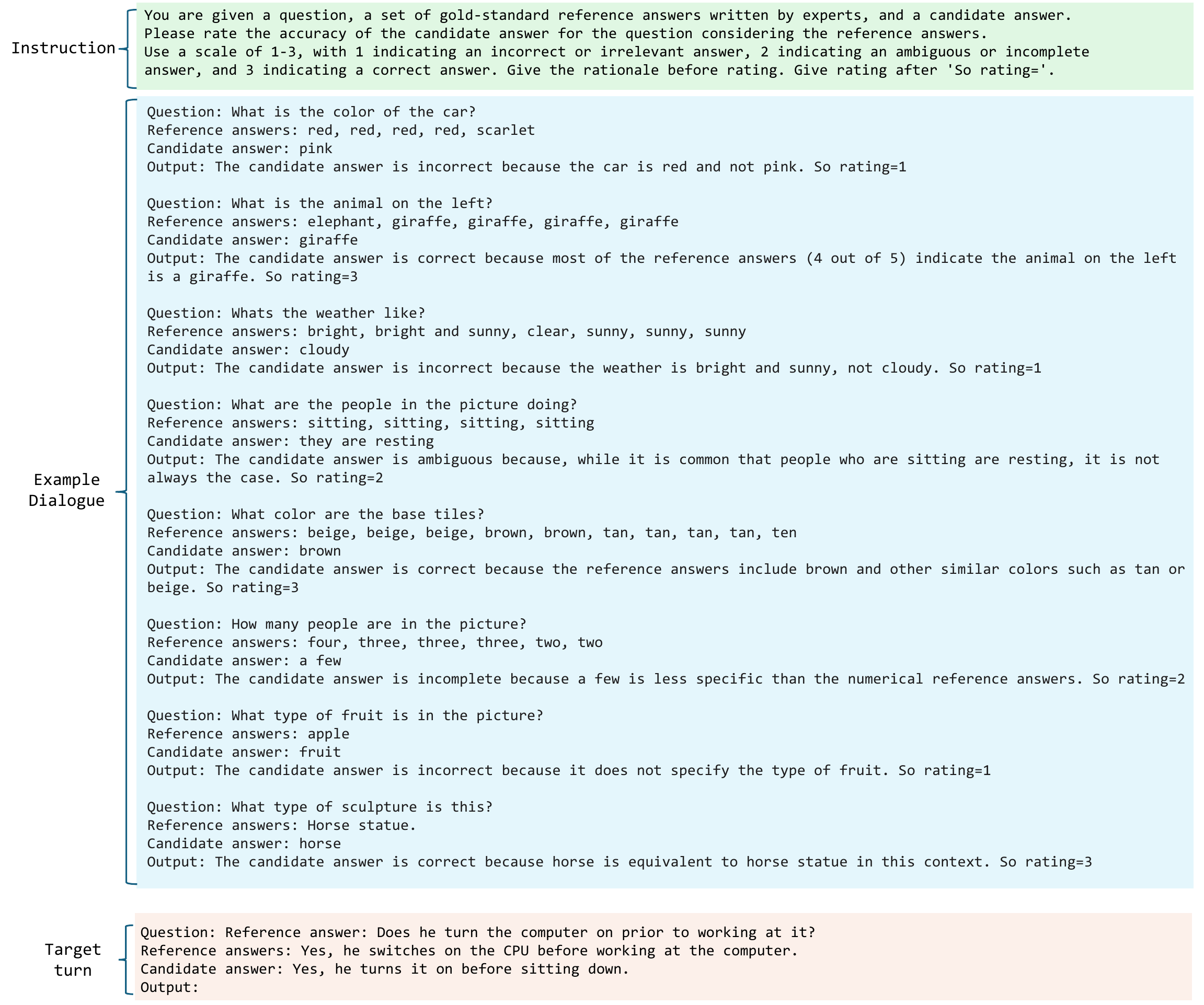}
    \caption{Whole prompt for LAVE.}
    \label{fig:prompt-lave}
\end{figure*}

Figure~\ref{fig:prompt-vdeval} shows the whole prompt used in the VDEval metric. For comparison, we show the prompt used in the LAVE metric in Figure~\ref{fig:prompt-lave}.
In VDEval, we include the entire dialogue history, which provides context from the previous question-answer pairs with evaluated relational and scores. Refined video summaries from scenario KGs are also added to the prompt, which provide a high-level overview of the video context for evaluation.
\subsection{Additional Results on VDAct}
\subsubsection{Generated Responses by Baseline Models}
Table \ref{tab:generated-responses} presents the actual generated responses of the first five turns for the dialogue by the best baseline when the model's parameters are frozen and fine-tuned with LoRA. Among these, only the responses generated at the 4th and 5th turns received a human evaluation score of 3, indicating that they were accurate. Meanwhile, for other dialogue turns, the generated responses are partially accurate or inaccurate.

\begin{table}[H]
    \fontsize{7}{9}\selectfont
    \centering
    \begin{tabular}{c|p{0.65\linewidth}|K{0.2cm}|K{0.2cm}|K{0.2cm}}
    \hline
    $t_i$ & & H & L & V \\ \hline
    1 & \multicolumn{4}{l}{\textbf{Question}: How does the man start his activities?} \\
    & \multicolumn{4}{p{0.75\linewidth}}{\textbf{Reference}: He starts by walking from the kitchen to the bathroom to get a towel so he can clean the TV.} \\ \cline{2-5}
    & \textbf{VL2}: The man starts his activities by walking towards the bed and looking out of the window. & 2 & 1 & 1 \\ \cline{2-5}
    & \textbf{VL2*}: He starts by walking into the bedroom and then returning to the living room. & 2 & 1 & 1 \\ \hline
    
    2 & \multicolumn{4}{l}{\textbf{Question}: How long does he spend cleaning the TV?} \\
    & \multicolumn{4}{p{0.75\linewidth}}{\textbf{Reference}: About 15 seconds.} \\ \cline{2-5}
    & \textbf{VL2}: He spends a long time cleaning the TV, as he is very meticulous and thorough in his cleaning. & 1 & $1^\dag$ & $1^\dag$ \\ \cline{2-5}
    & \textbf{VL2*}: He spends about 1 minute and 20 seconds cleaning the TV. & 1 & $1^\dag$ & $1^\dag$ \\ \hline

    3 & \multicolumn{4}{l}{\textbf{Question}: What did he do after that?} \\
    & \multicolumn{4}{p{0.75\linewidth}}{\textbf{Reference}: After that, he goes into his bedroom where he attempts to sit in a chair in the corner.} \\ \cline{2-5}
    & \textbf{VL2}: He goes back to the kitchen and gets a glass of water. & 1 & $1^\dag$ & $1^\dag$ \\ \cline{2-5}
    & \textbf{VL2*}: He went to the living room to do some research. & 1 & $1^\dag$ & $1^\dag$ \\ \hline

    4 & \multicolumn{4}{l}{\textbf{Question}: What happens when he attempts to sit?} \\
    & \multicolumn{4}{p{0.75\linewidth}}{\textbf{Reference}: He doesn't quite manage it, he awkwardly and somewhat creepily slides slowly out of the chair onto his hands and knees.} \\ \cline{2-5}
    & \textbf{VL2}: He falls over and then gets up and walks back to the kitchen. & 2 & 1 & $2^\dag$ \\ \cline{2-5}
    & \textbf{VL2*}: He falls over onto his hands and knees. & 3 & 2 & $3^\dag$ \\ \hline

    5 & \multicolumn{4}{l}{\textbf{Question}: Does he recover from his creepy fall and continue his activities?} \\
    & \multicolumn{4}{p{0.75\linewidth}}{\textbf{Reference}: Yes, somehow he does and he moves on to putting his slippers away.} \\ \cline{2-5}
    & \textbf{VL2}: Yes, he gets up and continues with his activities. & 3 & $3^\dag$ & 2 \\ \cline{2-5}
    & \textbf{VL2*}: Yes, he does. & 3 & 2 & 2 \\ \hline
    
    \end{tabular}
    \caption{Generated responses by a frozen and fine-tuned VideoLLaMA2 (denoted by VL and VL2*) for the first 5 turns of a dialogue ID 0073102, where a scenario includes Clean television3, Fall while sitting down, Put slippers in closet2, Drink water1, Pick up dirty dishes1. H, L, and V denote human rating, LAVE, and VDEval scores, respectively. $\dag$ represents a score that is correlated to human rating.}
    \label{tab:generated-responses}
\end{table}

\subsubsection{Performance on Different Question Types}
\begin{figure}[H]
    \centering
    \includegraphics[width=0.9\linewidth]{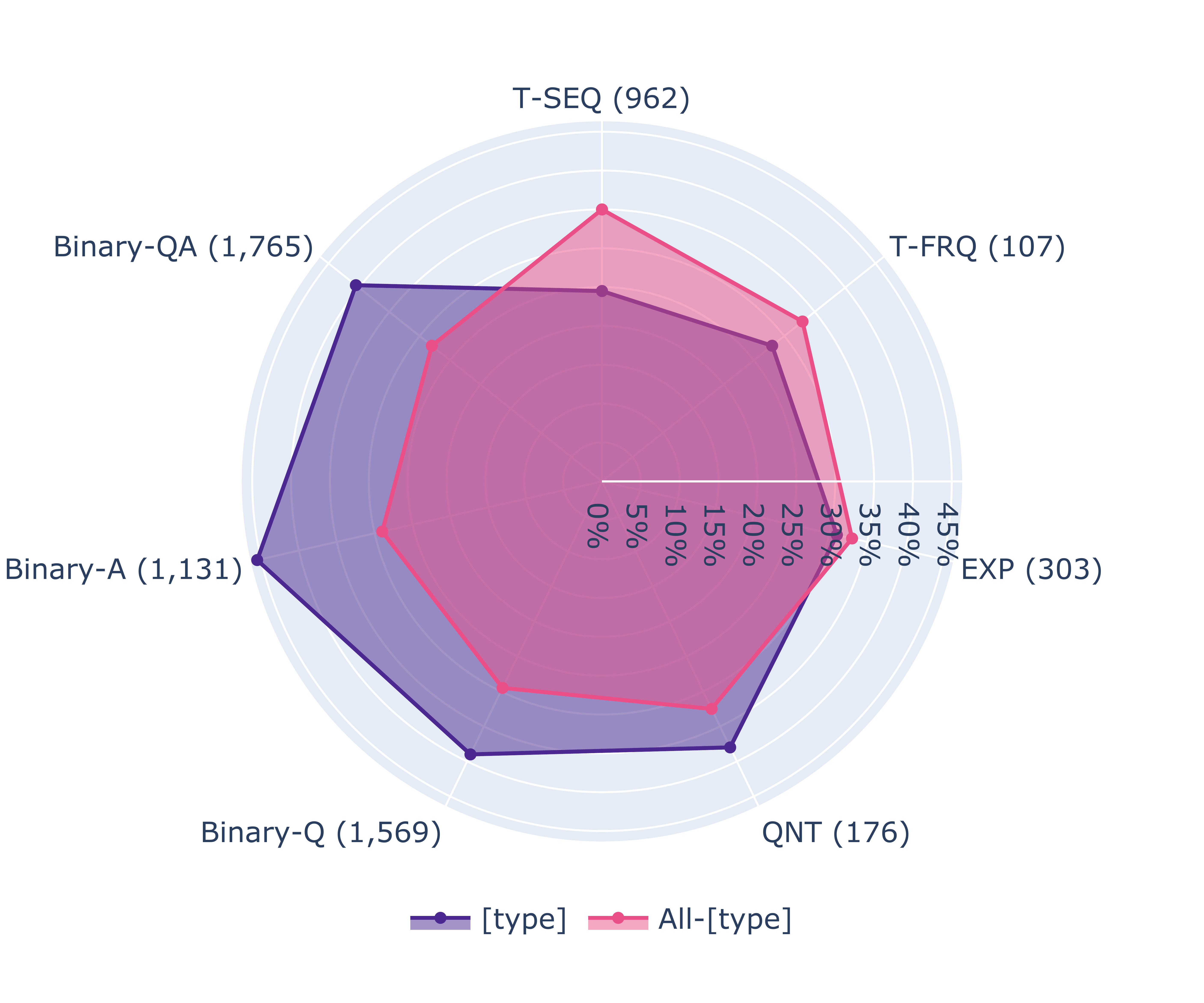}
    \caption{Performance of the best fine-tuned baseline with LoRA, VideoLLaMA2, using VDEval scores (reported in \% of 0-1 range conversion) on different question types in the VDAct test set. [type] and All-[type] indicate the question type and other questions that do not belong to the question type, respectively. The number in the parenthesis denotes the total number of questions for each type.}
    \label{fig:res-by-question-type}
\end{figure}

Figure \ref{fig:res-by-question-type} shows the performance of the fine-tuned VideoLLaMA2 with LoRA, evaluated using VDEval across various question types in the VDAct dataset. We categorized the questions based on specific keywords. For T-SEQ questions, we identified keywords such as ``first'', ``last'', ``before'', ``after'', and ``next''. T-FRQ questions were recognized by phrases like ``how long'' and ``how frequently''. EXP and QNT questions were distinguished by terms like ``how'' and ``how many/much''. Binary questions (Binary-Q) were identified by prefixes such as ``did he'' and ``is he'', while binary answers (Binary-A) were identified by ``yes/no'' responses. Binary-QA includes both Binary-Q and Binary-A.
While the model demonstrated a strong performance on some question types including binary (Binary-QA, Binary-Q, and Binary-A) and quantitative questions (QNT), the model performed poorly on question types such as temporal (T-SEQ and T-FRQ) and explanatory questions (EXP). The significant difference can be seen on T-SEQ where the average VDEval score for this type is lower than those achieved by other question types. This emphasizes the difficulty of generating accurate responses for temporal questions where more context reasoning and video understanding are required.

\subsubsection{Performance based on Question and Answer Lengths}
\begin{figure}[H]
    \centering
    \includegraphics[width=\linewidth]{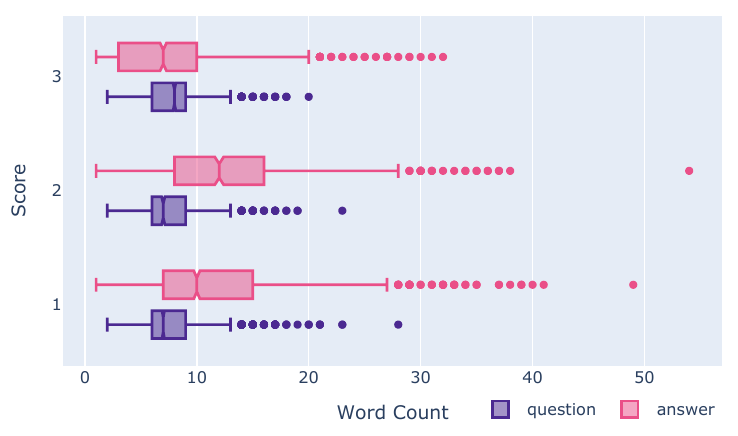}
    \caption{Performance of the best fine-tuned baseline with LoRA, VideoLLaMA2, using VDEval scores (reported in the original 1-3 range) on different question and answer lengths in the VDAct test set.}
    \label{fig:res-by-qa-length}
\end{figure}
We evaluated the performance of the fine-tuned VideoLLaMA2 with LoRA across different question and answer lengths, as depicted in Figure \ref{fig:res-by-qa-length}. The analysis reveals that question length does not significantly affect the model's performance, as VDEval scores for responses remained consistent across various question lengths. However, the model's performance declined for the generated responses where the reference answers are longer, as indicated by the lower VDEval scores (1 and 2) for these longer answers.
\subsection{Additional Results by VDEval}
\subsubsection{Comparison of Scores and Rationales by LAVE and VDEval}
Table~\ref{tab:comparison_lave_vdeval} shows a comparison of evaluation rationale and score from LAVE and VDEval on sets of questions, reference answers, and baseline model predictions. In the sample from dialogue ID 0037002, turn 3, LAVE gave a score of 1 to the answer of the baseline model by indicating that ``the reference answer does not specify that the book was taken from a shelf in the living room'' in the rationale. On the other hand, VDEval gave a score of 3 by considering not only the reference answer but also the video summary extracted from KGs. The rationale of VDEval indicated ``Both the summary and the reference confirm that the book was found on a bookshelf in the living room.''

In the sample from dialogue ID 0077002, turn 7, LAVE considers the completely incorrect answer as partially correct because LAVE does not refer to the dialogue history and summary for the question ``What else happens''. On the other hand, VDEval referred to the past dialogue turns and summary, correctly giving a score of 1 to the answer.

In the sample from dialogue ID 0005803, turn 4, the question ``What was this artwork of in the bedroom?'' is responded in the reference answer as ``It was the Mona Lisa.'' while the baseline model answered that it cannot be determined the content of the artwork. LAVE assigned a score of 1 because the model failed to identify the artwork, in contrast, VDEval assigned a score of 3, stating that the inability to identify the artwork is appropriate because the summary also lacks information to identify the content of the artwork. This example shows that the content of the video summary has a large impact on VDEval.

\begin{table*}[t]
    \fontsize{7}{8}\selectfont
    \centering
    \begin{tabular}{llp{0.75\linewidth}}
        \multicolumn{3}{l}{\textbf{Dialogue ID}: 0037002, \textbf{Turn}: 3}\\\hline
        \multicolumn{2}{l}{Question} &  Where did he get the book from? \\
        \multicolumn{2}{l}{Reference answer} & The bookcase. \\
        \multicolumn{2}{l}{Baseline prediction} & He got the book from the shelf in the living room.\\\hline
        \multirow{5}{*}{LAVE} & \multirow{4}{*}{Rationale} & The candidate answer is incorrect because it provides a more specific location (the shelf in the living room) rather than the general location given in the reference answer (the bookcase). While a shelf can be part of a bookcase, the reference answer does not specify that the book was taken from a shelf in the living room. Therefore, the candidate answer does not accurately reflect the information provided in the reference answer.\\\cline{2-3}
        & Score & 1\\\hline
        \multirow{9}{*}{VDEval} & \multirow{5}{*}{Summary} & The person moves from the kitchen to the living room, where he finds a book on a bookshelf. He grabs the book and sits on a sofa facing a TV, then reads the book. He later walks to the bathroom, grabs a towel, and returns to the living room to get a pillow. He uses the towel and shakes the pillow before sitting on the sofa again, where he also grabs a cellphone and texts. The person then goes to the kitchen to get and drink milk. Afterward, he walks to the bedroom, grabs a pillow from the bed, sits on the bed, then falls. \\\cline{2-3}
        & \multirow{3}{*}{Rationale} & The candidate answer accurately states that the man got the book from the shelf in the living room, which aligns with the reference answer that specifies the bookcase. Both the summary and the reference confirm that the book was found on a bookshelf in the living room. Therefore, the candidate answer is correct and fully addresses the question.\\\cline{2-3}
        & Score & 3\\\hline
        \\
        \multicolumn{3}{l}{\textbf{Dialogue ID}: 0077002, \textbf{Turn}: 7}\\\hline
        \multicolumn{2}{l}{Question} &  What else happens?\\
        \multicolumn{2}{l}{Reference answer} & He puts some dirty water glasses in the sink. \\
        \multicolumn{2}{l}{Baseline prediction} & He then goes to the bathroom and turns on the faucet.\\\hline
        \multirow{6}{*}{LAVE} & \multirow{5}{*}{Rationale} & The candidate answer provides additional information about what the subject does after the initial action of putting dirty water glasses in the sink. However, it does not directly address the specific action mentioned in the reference answer. The reference answer focuses on the action of putting dirty glasses in the sink, while the candidate answer introduces a new action (going to the bathroom and turning on the faucet) that is not mentioned in the reference. Therefore, the candidate answer is somewhat relevant but does not accurately reflect what else happens according to the reference.\\\cline{2-3}
        & Score & 2\\\hline
        \multirow{13}{*}{VDEval} & \multirow{6}{*}{Summary} & The person is in the kitchen and walks to the living room, where he sits on the sofa. After standing up, he walks to the TV and then to the kitchen, where he falls. He approaches the sink and the washing sponge on the counter, picks it up, and wipes the sink. He operates the faucet by turning it on and off before wiping the sink again. The person then grabs a wineglass from the table, pours wine into two glass, and places them down. He then walks to the bedroom and the bathroom, where he retrieves a toothbrush and toothpaste from the counter. After squeezing the toothpaste, he turns on the faucet to soak the toothbrush before switching it off. \\\cline{2-3}
        & \multirow{2}{*}{Previous turns} & Turn=5 Q: What's next on the list of activities? A: He cleans the sink in the kitchen.\\
        & & Turn=6 Q: How does he do this? A: With a sponge and some tap water.\\\cline{2-3}
        & \multirow{4}{*}{Rationale} & The candidate answer does not address the question about what else happens after the previous activities. Instead, it introduces a new action that is not mentioned in the reference answer or the summary. The reference answer specifies that he puts dirty water glasses in the sink, which is a key detail that the candidate answer completely omits. Therefore, the candidate answer is incorrect and does not align with the reference answer.\\\cline{2-3}
        & Score & 1\\\hline
        \\
        \multicolumn{3}{l}{\textbf{Dialogue ID}: 0005803, \textbf{Turn}: 4}\\\hline
        \multicolumn{2}{l}{Question} & What was this artwork of in the bedroom? \\
        \multicolumn{2}{l}{Reference answer} & It was the Mona Lisa. \\
        \multicolumn{2}{l}{Baseline prediction} & I'm sorry, I cannot determine the content of the artwork as it was not specified.\\\hline
        \multirow{4}{*}{LAVE} & \multirow{3}{*}{Rationale} & The candidate answer is incorrect because it fails to acknowledge the specific reference answer that identifies the artwork as the Mona Lisa. Instead of providing a definitive response, the candidate answer expresses uncertainty, which does not align with the clear information given in the reference.\\\cline{2-3}
        & Score & 1\\\hline
        \multirow{9}{*}{VDEval} & \multirow{4}{*}{Summary} & The person is in the bedroom and walks to the kitchen. In the kitchen, he approaches a wall picture frame, looks at it, and then notices the wine on the kitchen table. He grabs the wine and places it on the table. Next, he grabs a water glass from the table, walks to the sink, and pours it. Afterward, he returns to the bedroom where he finds a pillow on the bed. He grabs the pillow, which is facing the computer, then walks to the bed, sits down, stands, and falls. \\\cline{2-3}
        & \multirow{4}{*}{Rationale} & The candidate answer correctly states that it cannot determine the content of the artwork because the summary does not provide that information. This aligns with the reference answer, which specifies the artwork in the bedroom as the Mona Lisa, but the candidate does not have that information available. Therefore, the candidate answer is accurate in its acknowledgment of the lack of detail regarding the artwork's content.\\\cline{2-3}
        & Score & 3\\\hline

    \end{tabular}
    \caption{A comparison of LAVE and VDEval's rationale and score for the sets of questions, reference answers, and predictions by frozen VideoLLaMA2. In the second example, only two previous dialogue turns are included due to the space limitation.}
    \label{tab:comparison_lave_vdeval}
\end{table*}

\clearpage

\section{Reproducibility Checklist}
This paper: (Our responses are in bold and underlined.)

\begin{itemize}
    \item Includes a conceptual outline and/or pseudocode description of AI methods introduced: \textbf{\underline{yes}}
    \item Clearly delineates statements that are opinions, hypothesis, and speculation from objective facts and results: \textbf{\underline{yes}}
    \item Provides well marked pedagogical references for less-familiare readers to gain background necessary to replicate the paper: \textbf{\underline{yes}}

    \item Makes theoretical contributions: \textbf{\underline{yes}}
    \begin{itemize}
        \item All assumptions and restrictions are stated clearly and formally. (\textbf{\underline{yes}}/partial/no)
        \item All novel claims are stated formally (e.g., in theorem statements). (\textbf{\underline{yes}}/partial/no)
        \item Proofs of all novel claims are included. (\textbf{\underline{yes}}/partial/no)
        \item Proof sketches or intuitions are given for complex and/or novel results. (\textbf{\underline{yes}}/partial/no)
        \item Appropriate citations to theoretical tools used are given. (\textbf{\underline{yes}}/partial/no)
        \item All theoretical claims are demonstrated empirically to hold. (\textbf{\underline{yes}}/partial/no/NA)
        \item All experimental code used to eliminate or disprove claims is included. (yes/\textbf{\underline{no}}/NA)
    \end{itemize}

    \item Relies on one or more datasets: \textbf{\underline{yes}}
    \begin{itemize}
        \item A motivation is given for why the experiments are conducted on the selected datasets (\textbf{\underline{yes}}/partial/no/NA)
        \item All novel datasets introduced in this paper are included in a data appendix. (\textbf{\underline{yes}}/partial/no/NA)
        \item All novel datasets introduced in this paper will be made publicly available upon publication of the paper with a license that allows free usage for research purposes. (\textbf{\underline{yes}}/partial/no/NA)
        \item All datasets drawn from the existing literature (potentially including authors’ own previously published work) are accompanied by appropriate citations. (\textbf{\underline{yes}}/no/NA)
        \item All datasets drawn from the existing literature (potentially including authors’ own previously published work) are publicly available. (\textbf{\underline{yes}}/partial/no/NA)
        \item All datasets that are not publicly available are described in detail, with explanation why publicly available alternatives are not scientifically satisficing. (yes/partial/no/\textbf{\underline{NA}})
    \end{itemize}
    \item Includes computational experiments: \textbf{\underline{yes}}
    \begin{itemize}
        \item Any code required for pre-processing data is included in the appendix. (yes/partial/\textbf{\underline{no}}).
        \item All source code required for conducting and analyzing the experiments is included in a code appendix. (yes/partial/\textbf{\underline{no}})
        \item All source code required for conducting and analyzing the experiments will be made publicly available upon publication of the paper with a license that allows free usage for research purposes. (\textbf{\underline{yes}}/partial/no)
        \item All source code implementing new methods have comments detailing the implementation, with references to the paper where each step comes from (yes/partial/\textbf{\underline{no}})
        \item If an algorithm depends on randomness, then the method used for setting seeds is described in a way sufficient to allow replication of results. (\textbf{\underline{yes}}/partial/no/NA)
        \item This paper specifies the computing infrastructure used for running experiments (hardware and software), including GPU/CPU models; amount of memory; operating system; names and versions of relevant software libraries and frameworks. (\textbf{\underline{yes}}/partial/no)
        \item This paper formally describes evaluation metrics used and explains the motivation for choosing these metrics. (\textbf{\underline{yes}}/partial/no)
        \item This paper states the number of algorithm runs used to compute each reported result. (\textbf{\underline{yes}}/no)
        \item Analysis of experiments goes beyond single-dimensional summaries of performance (e.g., average; median) to include measures of variation, confidence, or other distributional information. (\textbf{\underline{yes}}/no)
        \item The significance of any improvement or decrease in performance is judged using appropriate statistical tests (e.g., Wilcoxon signed-rank). (\textbf{\underline{yes}}/partial/no)
        \item This paper lists all final (hyper-)parameters used for each model/algorithm in the paper’s experiments. (\textbf{\underline{yes}}/partial/no/NA)
        \item This paper states the number and range of values tried per (hyper-) parameter during development of the paper, along with the criterion used for selecting the final parameter setting. (\textbf{\underline{yes}}/partial/no/NA)
    \end{itemize}
\end{itemize}

\end{document}